\documentclass[a4paper, 10pt, conference]{ieeeconf}      

\IEEEoverridecommandlockouts                              

\usepackage{graphics} 
\usepackage{epsfig} 
\usepackage{float} 

\usepackage{mathptmx} 
\usepackage{times} 
\usepackage{amsmath} 
\usepackage{amssymb}  

\usepackage{dsfont}
\usepackage{soul}
\usepackage[normalem]{ulem}
\usepackage[utf8]{inputenc}
\usepackage[small]{caption}
\usepackage{graphicx}
\usepackage{amsmath}
\usepackage{booktabs}
\usepackage{algorithm}
\usepackage[noend]{algorithmic}
\usepackage{amsfonts}
\usepackage{blindtext}
\usepackage{subfig}
\graphicspath{{Figures//}{Figures/si-figures//}}
\DeclareGraphicsExtensions{.eps,.pdf,.png,.jpg,.tif,.tiff,.ps}

\usepackage{caption}
\usepackage{tabu}

\makeatletter
\let\NAT@parse\undefined
\makeatother
\usepackage[hidelinks]{hyperref}  
\usepackage{url}
\usepackage[nameinlink,capitalize]{cleveref}
\usepackage{soul}
\usepackage[colorinlistoftodos]{todonotes}

\definecolor{navy}{rgb}{0.1, 0.1, 0.8}
\definecolor{gray}{rgb}{0.6, 0.6, 0.6}
\definecolor{myblue}{rgb}{.8, .8, 1}
\definecolor{olive}{rgb}{0.1, 0.5, 0.1}
\newcommand{\eat}[1]{}

\newcommand{\revA}[1]{{\color{black}{#1}}}

\newcommand{\verify}[1]{{\color{black}{#1}}}

\usepackage{xspace}
\newcommand{\ti}{\textsc{Ti}\xspace}

\newcommand{\citet}[1]{\citeauthor{#1}~\shortcite{#1}}
\newcommand{\citep}{\cite}

\newcommand{\titlename}{Motorway Traffic Flow Prediction using Advanced Deep Learning}

\title{\LARGE \bf \titlename}

\author{Adriana-Simona Mihaita$^{1}$, Haowen Li$^{2}$, Zongyang He$^{2}$, Marian-Andrei Rizoiu$^{1}$ 
\thanks{$^{1}$Adriana-Simona Mihaita and Marian-Andrei Rizoiu are with the University of Technology in Sydney, Faculty of Engineering and IT, School of Computer Science: 81 Broadway Ultimo, NSW, Australia. 
Dr Mihaita \revA{is also affiliated with CSIRO's Data61}.
Corresponding authors contact: {\tt\small adriana-simona.mihaita@uts.edu.au}, {\tt\small marian-andrei.rizoiu@uts.edu.au}}.  
\thanks{$^{2}$ Haowen Li and Zongyang He are with the Australian National University, Canberra, ACT, Australia. Emails: u6342101@anu.edu.au, u6342051@anu.edu.au.} 
}

\begin{document}

\maketitle
\thispagestyle{empty}
\pagestyle{empty}

\begin{abstract}
Congestion prediction represents a major priority for traffic management centres \revA{around the world} to ensure timely incident response handling. 
\revA{The increasing amounts of generated traffic data have been used to train machine learning predictors for traffic, however this is a challenging task} due to inter-dependencies of traffic flow both in time and space. 
Recently, deep learning techniques have shown significant \revA{prediction} improvements over traditional models, however open questions remain around their applicability, accuracy and parameter tuning. 
This paper proposes an advanced deep learning framework for simultaneously predicting the traffic flow on a large number of monitoring stations along a highly circulated motorway in Sydney, Australia, including exit and entry loop count stations, and over varying training and prediction time horizons.
The spatial and temporal features extracted from the 36.34 million data points are used in various deep learning architectures that exploit their spatial structure (convolutional neuronal networks), their temporal dynamics (recurrent neuronal networks), or both through a hybrid spatio-temporal modelling (CNN-LSTM).
We show that our deep learning models consistently outperform traditional methods, and we conduct a comparative analysis of the optimal time horizon of historical data required to predict traffic flow at different time points in the future.
\end{abstract}



%
\section{INTRODUCTION}\label{S1_Intro}

\nocite{appendix}
Traffic congestion represents one of the sensitive points of many traffic management centres around the world, which need to ensure that travel times remain within regular patterns, and that incidents are cleared in due time on a daily basis. 
Predicting the dynamics of the traffic congestion within the next 30min to one hour represents a high priority for real-time traffic operations. 
The main advantage of using advanced traffic flow prediction \revA{techniques} lies in \revA{their} ability to quickly adapt to stochastic incidents, and to predict their impact starting from only incipient measurements. 
The increasing amount of traffic data generated by intelligent transport systems led to the development of multiple data-driven approaches and prediction models.
However, there are several open questions concerning traffic flow prediction:
a) how to efficiently predict road traffic congestion using extensive data-driven techniques which can adapt to real-time big-data sets?, 
b) what are the best techniques that can capture the spatial-temporal correlations arising in complex traffic networks? and 
c) why are some models efficient for short-term traffic prediction, but not for long-term prediction? 

The initial approaches developed to answer these questions have been focused on parametric and non-parametric models for short-term traffic predictions.
The parametric approaches were typically based on time-series analysis, Kalman models, etc. For example, the ARIMA model has been widely applied for traffic flow prediction (\cite{VANDERVOORT1996307, dKamarianakis2003}) due to its simplicity and good performance in forecasting linear and stationary time-series. 
Further extensions of ARIMA have been proposed to account for seasonal features (SARIMA \cite{quteprints63176}), and for additional explanatory variables (ARIMAX \cite{Williams2001}). 
The effectiveness of parametric models can be affected by the traffic stochasticity, and by the occurence of disruptive events.
As a result, non-parametric models have seen an increasing popularity, and among them we cite: k-nearest neighbours \cite{Chang2012}, support vector regressions \cite{Jeong2013}, artificial neural networks \cite{KARLAFTIS2011}, and Gaussian Processes \cite{IDe2009}.

Recently Deep Learning (DL) methods have emerged as popular non-parametric alternative approaches for short-term predictions, with various models being proposed and tested in different set-ups. 
Two major literature reviews on DL models can be found in \cite{Ali2018}, and \cite{Nguyen2018}.
These debate how different DL models can be adapted for traffic flow prediction, and why the spatial and temporal correlation in traffic congestion propagation makes the application of such models difficult. 
More recently, recurrent neural networks (RNNs) -- such as the long short-term memory model (LSTM) -- have been designed to learn from sequences of data, and to capture long-term temporal patterns.
LSTM was applied to traffic flow data~\cite{TIAN2018297}, either alone of jointly with convolutional neuronal networks (CNN) \cite{Nguyen2019,Wu2019} in an attempt to capture the spatial road network information.
Among the difficulties of deploying such models are their often complicated structure, the choice of parameters (such as the number of neurons or the non-linear functions), and the fact that neural networks have been long time regarded as ``black-boxes'' \cite{ZHANG201465}.
These difficulties are starting to ease due to the emergence of integrated modelling and fitting frameworks, such as TensorFlow~\cite{tensorflow2015} and PyTorch~\cite{paszke2017automatic}.

%
\begin{figure*}[h]
	\centering
	\includegraphics[width=0.99\textwidth]{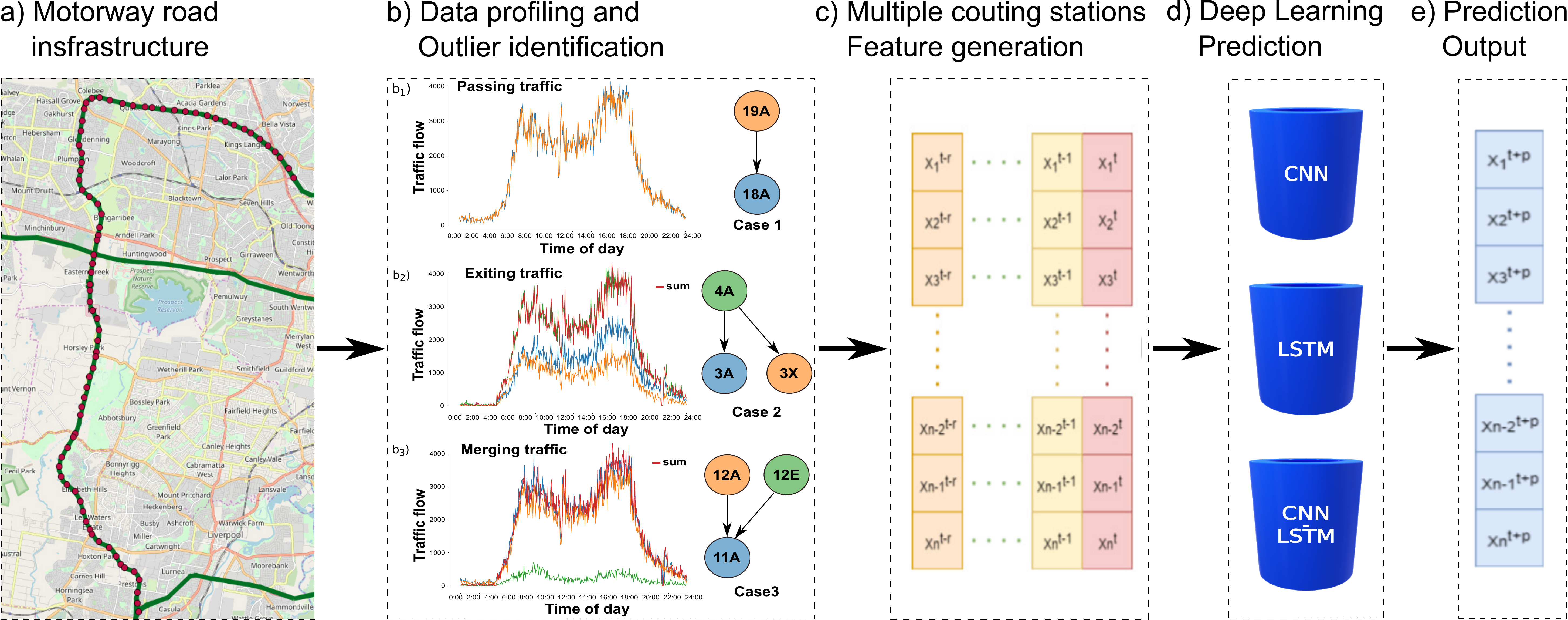}
	\caption{
		Schema of the proposed DL methodology for the motorway flow prediction.
	} 
	\label{Fig_0_Methodology}
\end{figure*}

There are several open questions relating to the usage of DL methods for traffic flow prediction, still not addressed in the literature.
The first question relates to the scalability of such methods.
The majority of existing studies concentrate on one or several stations, or over short periods of time~\cite{Tan2018}.
It is therefore unclear how DL models behave at the level of an entire motorway, or for large
datasets with complicated road structures. 
We address this question by
constructing deep learning models capable of predicting the real-time traffic flow along an entire motorway.
Our dataset spans over an year, and it spreads across 208 stations along 48kms of road network in Sydney, Australia. 

The second open question relates to the relationship between the training and the prediction horizons.
The majority of existing work performs traffic flow prediction for future time horizons of 2-3 time intervals~\cite{POLSON20171}
however they usually fix the past horizon to 30 or 40 minutes.
This arbitrary choice can affect the prediction performances, and impact the model selection. 
In our study we propose a sensitivity analysis between various past and future horizons for each of the deep learning models under investigation, and we showcase the best set-up for each model.
We find that for LSTM there is a limit in the past horizon beyond which the accuracy stops improving, and that CNN performances are actually hurt by using too much data from the past.

The third open question is about deploying hybrid deep learning models that combine both spatial and temporal modelling.
While some research has shown that hybrid DL architectures can improve performances in specific circumstances \cite{Tan2018,Wang2016TrafficSP}, other studies still debate the necessity and overhaul of fine-tuning such models. 
In this paper, we implement a hybrid CNN-LSTM model and we find that it under-performs the LSTM model, with its performances fluctuating significantly with respect to the future time horizon.

The object of this paper is to construct an advanced deep learning framework for predicting the traffic flow, which can be used to perform model comparison under varying prediction conditions and which can serve as a benchmark for future predictive work. 
We apply our models on flow counting stations along a long motorway in Australia, and we consider both the spatial structure of the datasets, as well as the historical flow during our one year long dataset. 
We first present the methodology for constructing various deep learning models such as CNN, LSTM and the hybrid architecture CNN-LSTM, and we compare them against typically employed approaches, such as ARIMA, individual station regressors, and the average historical traffic flow.
We follow with a sensitivity analysis of historical versus prediction time horizons, and their impact on model performance.
We end with the comparative analysis of the prediction performances of the models, and a model choice discussion.


%
\section{Methodology}\label{S2_Method}

In this section we present the proposed deep learning methodology for predicting the traffic flow along motorways.
\cref{Fig_0_Methodology} presents the proposed methodological framework, which consists of four steps: network identification (detailed in \cref{subsec:network-ident}), data profiling (\cref{subsec:profiling}), feature generation (\cref{subsec:feature-generation}) and DL model development and traffic prediction (\cref{subsec:models}).

\subsection{\textbf{Network identification and data set preparation}} 
\label{subsec:network-ident}

We first gather the spatial information with regards to the placement of traffic monitoring stations along the motorway segments, the road network geometry file and the temporal information in the form of traffic flow recorded at each time step (3min time-intervals in our case).
The obtained dataset is described in \cref{subsec:dataset}.

\subsection{ \textbf{Data profiling and outlier identification}} 
\label{subsec:profiling}
This step is necessary for building regular traffic patterns depending on the type of day, time-of-day, etc. 
\cref{Fig_0_Methodology} showcases 3 different possible cases of station spatial structure,
which are automatically checked for data accuracy and motorway structure consistency in both time and space. 
Let $T_f(\texttt{18A})$ and $T_f(\texttt{19A})$ be the traffic flows registered at the stations $\texttt{18A}$, and $\texttt{19A}$ respectively. 
If no exit or entry is recorded between two consecutive stations (see \cref{Fig_0_Methodology}-$b_1$), we assume that the two flow patterns should match, and we check for consistency as  $T_f(\texttt{18A}) = T_f(\texttt{19A}) \pm \epsilon$ ($\epsilon$ accounts for the inherent detector equipment error). 
When exit loops exist (see \cref{Fig_0_Methodology}-$b_2$), the module verifies that the sum of the flow recorded at the downstream stations ($\texttt{3A}$ and $\texttt{3X}$) matches the traffic flow recorded at the closest upstream station: $T_f(\texttt{4A})=T_f(\texttt{3A})+T_f(\texttt{3X}) \pm \epsilon$. 
Lastly, in case of entry loops (see \cref{Fig_0_Methodology}-$b_3$), the module checks that the sum of downstream traffic flow is the sum of upstream flows: $T_f(\texttt{11A})=T_f(\texttt{12A})+T_f(\texttt{12E})\pm \epsilon$. 
The data processing step also builds the daily flow patterns for all the stations along the entire motorway.
These are further used in the deep learning methodology, and to identify missing data and abnormal traffic flow due to traffic disruptions. 
This is further discussed in \cref{3_Case_study}. 
 
\subsection{\textbf{Feature construction}}
\label{subsec:feature-generation}

The traffic flow is recorded as time series associated with each monitoring station (including entries and exits).
It is processed and transformed into sequential matrices, which we denote as $\boldsymbol{X^t}$ and which are the input of our DL models: 
\begin{equation}\label{Eq1}
\boldsymbol{X^t}= 
\begin{bmatrix}
\boldsymbol{\vec{X_1}^t} \\
\boldsymbol{\vec{X_2}^t}\\
...\\
\boldsymbol{\vec{X_N}^t}
\end{bmatrix}
=
\begin{bmatrix}
x_1^{t-R+1}  & ... & x_1^{t-1}   & x_1^t \\
x_2^{t-R+1}  &... & x_2^{t-1}    & x_2^t \\
... 		 & ...        &...  & ...\\
x_N^{t-R+1}  & ... & x_N^{t-1}   & x_N^t \\
\end{bmatrix}
\end{equation}
%
%
where $N$ is the total number of monitoring stations along the motorway;
$x_j^t, j=\{1,...N\}$ is the traffic flow registered at station $j$ at the time point $t$;
$R$ is the number of historical points used to train the models; 
$t-R$ will be often referred to as the training horizon or a ``past time-window'' and $\boldsymbol{\vec{X_i}^t}$ the past horizon (training) vector for each station. 
 
Our prediction target is $\boldsymbol{\hat{X}^{t+P}} = [\hat{X}_{1}^{t+P}, \hat{X}_{2}^{t+P},... ,\hat{X}_{n}^{t+P}]^T$ where $P$ denotes the ``prediction horizon'' (how far in the future we want to make the prediction) and $\hat{X}_{j}^{t+P} = \begin{bmatrix} x_j^{t+1} & x_j^{t+2}  & ..&  x_j^{t+P} \end{bmatrix} $ is the predicted traffic flow at the $j^{th}$ station over the prediction horizon. A summary of all the notations used in this paper is provided in \cref{tab:notations}.
%
\begin{table}[btp]
	\caption{Summary of notations.}
	\small
	\centering
	\begin{tabular}{cp{6.5cm}}
	\toprule
		Notation & Interpretation \\ 
		\midrule
		
		$N$ & the total number of stations used in this study ($N = 208$ comprised of $104$ in each direction). \\
		\ti & a 3-min \emph{Time Interval}; the time is discretized into 3-minute time intervals (480 \ti per day). \\
		$R$ & the length of the time window in the past; the number of \ti used as historic information.\\
		$P$ & future prediction horizon; predictions will be made for the $P^{th}$ \ti in the future. \\
		$x^{t-i}_j$ & the traffic flow of station $j$ at $\ti=t-i, i\in\{0,...R\}$.\\
				
		$\boldsymbol{\vec{S}^{t-i}}$  & the traffic flow for ALL stations at $\ti=t-i,i\in\{0,...R\}$; an $N$-dimensional column vector $\vec{S}^i$ = $[x^{t-i}_1$, $x^{t-i}_2$, ..., $x^{t-i}_N]^T$. \\
		$\boldsymbol{X^t}$ & the observed traffic flow, for all stations for the past $R$ \ti, an $R \times N$ matrix (see \cref{Eq1}). \\

		$RMSE$ & the Root Mean Square Error evaluation metric; $RMSE = \sqrt{\frac{1}{N} \sum_{j=1}^{N}\left(f_j-\hat{f_j} \right)^2 }$ \\
		$ReLU(x)$ & The ReLU function; $ReLU(x) = max(x,0)$ \\ \bottomrule
	\end{tabular}
	\label{tab:notations}
\end{table}

\subsection{\textbf{Deep learning model development}}
\label{subsec:models}

Deep learning is usually used to learn high dimensional functions via sequences of semi-affine non-linear transformations, and it has been shown effective general function learners~\cite{POLSON20171}.
A deep learning predictor is capable of addressing the non-linearity in the datasets, and of finding the spatio-temporal relations between features. 
We implement various DL models in the current modelling framework and we apply them either individually -- Convolutional Neural Networks (CNN), Long Short-Term memory networks (LSTM), back-propagation neuronal networks (BPNN) -- or in hybrid structures as an advanced deep learning architecture. 
We compare their performance to other parametric and baseline models in \cref{subsec:performance}. 
In the following we \revA{detail each of the DL} models and \revA{we} provide \revA{their} internal architecture setup. 

\paragraph{Back-propagation Neuronal Networks (BPNN)}

BPNN is a typical feed-forward type network which learns the relation between inputs and outputs without an explicit mapping of the information, and using a gradient descent optimisation method. 
Such models have been successfully applied for highway traffic incident detection \cite{CHENG2010482}, and also for tourist volume forecasting in Baidu \cite{LI2018116}.  


The topology of BPNN usually includes an input layer, one or multiple hidden layers, and an output layer.   
In our DL work, we developed a BPNN model which consists of two fully-connected layers. 
The input of first layer is the historical information of all stations, and the last layer's output is the prediction of the traffic flow across all monitoring stations. 
In this work, BPNN is mainly used as a lower-bound DL performance measure, and it serves to assess the performance gains obtained when implementing the more complex models detailed here below. 
 

\paragraph{Convolutional Neural Networks (CNN)}
In order to take into consideration the spatial features of the traffic flow data, in which the individual counting stations reflect the propagation of flow in, out and along the motorway (see \cref{Fig_0_Methodology}-b)), we employ a CNN model on the traffic dataset.
CNNs are bio-inspired models which have been widely applied for processing images, speech and time series \cite{AAAI1714501}.
The main feature of CNN is the convolution operator, which slides on a two-dimensional surface, smoothing it and extracting higher level abstraction.
In image processing, multiple convolution operations are subsequently applied to increase the abstraction of the information.
The input of a CNN can include both a spatial dimension and a temporal dimension, and the convolution operator can be applied in either one dimension (1D) or both dimensions (2D).
In various advanced DL architectures, CNN models usually contain various convolutional layers with non-linear activation functions applied in between. 

\begin{figure}[tbp]
	\centering
	\includegraphics[width=0.49\textwidth]{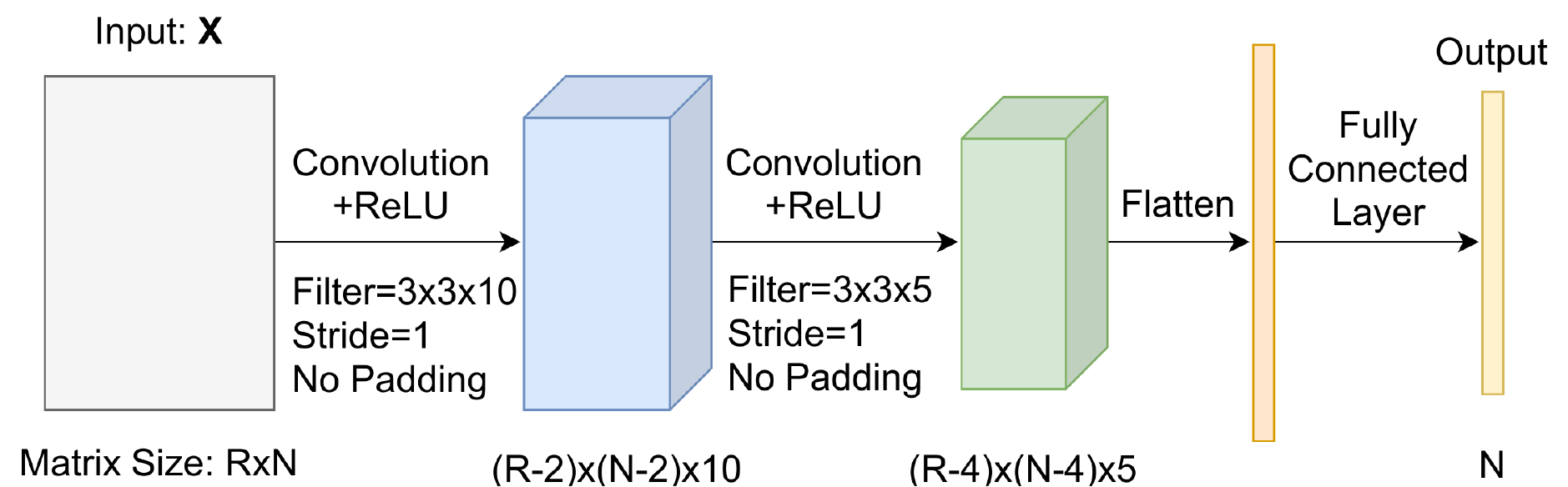}
	\caption{CNN model for traffic flow prediction.}
	\label{Fig2_CNN}
\end{figure}

For our study, we construct a fully-connected CNN structure as presented in \cref{Fig2_CNN}, which accepts as input the RxN feature matrix $\boldsymbol{X^t}$ defined in \cref{Eq1}. 
This 2-dimensional input is passed through two convolutional layers and two Rectified linear activation unit (ReLU) functions, before finally being flattened as a 1-D vector and sent through a Fully Connected Layer which outputs the final results. 
No pooling layers are employed in our model; although, in general, pooling layers may increase the speed of training and achieve better performance on high dimensional image recognition tasks, the spatial dimension of our data set is smaller and the information it contains is not redundant like in images. 
For advanced DL architectures, the ReLU function are more popular than the traditional sigmoid/tanh functions because: a) they are more computationally efficient, b) they can help avoid the exploding and vanishing gradient problems and c) they tend to show better performance in practice \cite{NIPS2012_4824}.  

\paragraph{Long Short-Term Memory Networks (LSTM)}
The temporal features of the traffic flow have a different representation than  the spatial one.
While the traffic flow in one station can be localy determined by the neighbouring stations (see the three interconnected cases depicted in \cref{Fig_0_Methodology}-b)), various external events (e.g., accidents or weather conditions) can cause traffic congestion in the downstream part of the motorway, which propagates to the upstream stations and eventually affects the travel time along the entire motorway. 
In order to model these long-term dependencies in traffic flow, the long short-term memory units (LSTM) have been introduced for achieving a balance between immediate (short term) inputs and historical (long term) trends
(see \cite{Kang2017,YANG2019320}). 
\begin{figure}[h!]
	\centering
	\includegraphics[width=0.49\textwidth]{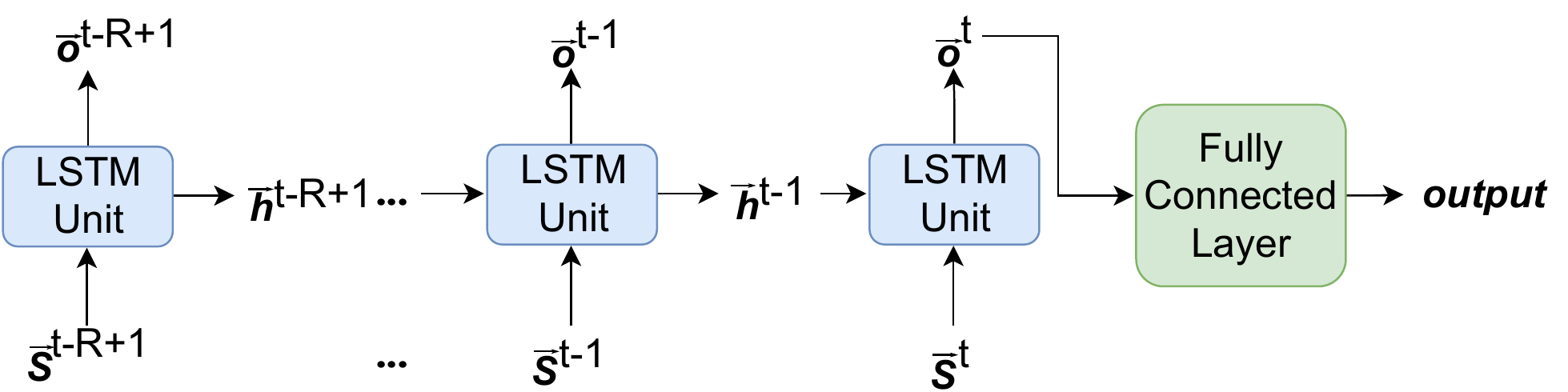}
	\caption{LSTM model for traffic flow prediction.}
	\label{Fig3_LSTM}
	\vspace{-.1cm}
\end{figure}

\cref{Fig3_LSTM} showcases the structure of the LSTM model that we develop in this work for the traffic flow prediction.
An LSTM \emph{unit} is typically comprised of an input layer, a hidden layer (which acts as a memory block containing an input gate, a forget gate and an output gate) and an output layer. 
The LSTM \emph{model} is a sequence of LSTM units, in which the output of one unit is consumed as input by the following unit.
The output of the last unit feeds into a fully connected layer which makes the the final prediction. 

In our application, the input feature matrix $\boldsymbol{X^t}$ is split into various flow vectors, one for each counting station (denoted as $\boldsymbol{\vec{S}^{t-i}}$). 
For each time step from the past horizon $\{t-R+1,.. t-1, t\}$, an LSTM unit accepts the vector $\boldsymbol{\vec{S}^{t-i}}$ as input and outputs a hidden state vector $h^i$ and an output vector $o^i$, of equal lengths. 
The hidden state $h^i$ is passed at the next unit $\{t-i+1\}$. 
The last output vector $o^t$ is connected to the fully-connected layer to get the final result. 
For our work, we only used only one hidden layer per unit and no drop-out layers, as we have a limited number of counting stations. 
For a more complex road network, this LSTM structure can be further extended.

%
%
%
%
%
%
%

\paragraph{Hybrid CNN-LSTM prediction}

We develop the hybrid model as
a combination of spatial and temporal processing, modelled by 
connecting the output of CNN to the input of each LSTM unit.
The intuition is that the structure of LSTM would learn the temporal patterns, while the structure of CNN can learn the location features. 
The final prediction is made using a Fully Connected layer, just like for LSTM. 
Several previous research works have employed the hybrid model and they report contradictory results: some argue that it improves the prediction accuracy, while others indicate the contrary~\cite{Nguyen2018}. 
\begin{figure}[h!]
	\centering
	\includegraphics[width=0.49\textwidth]{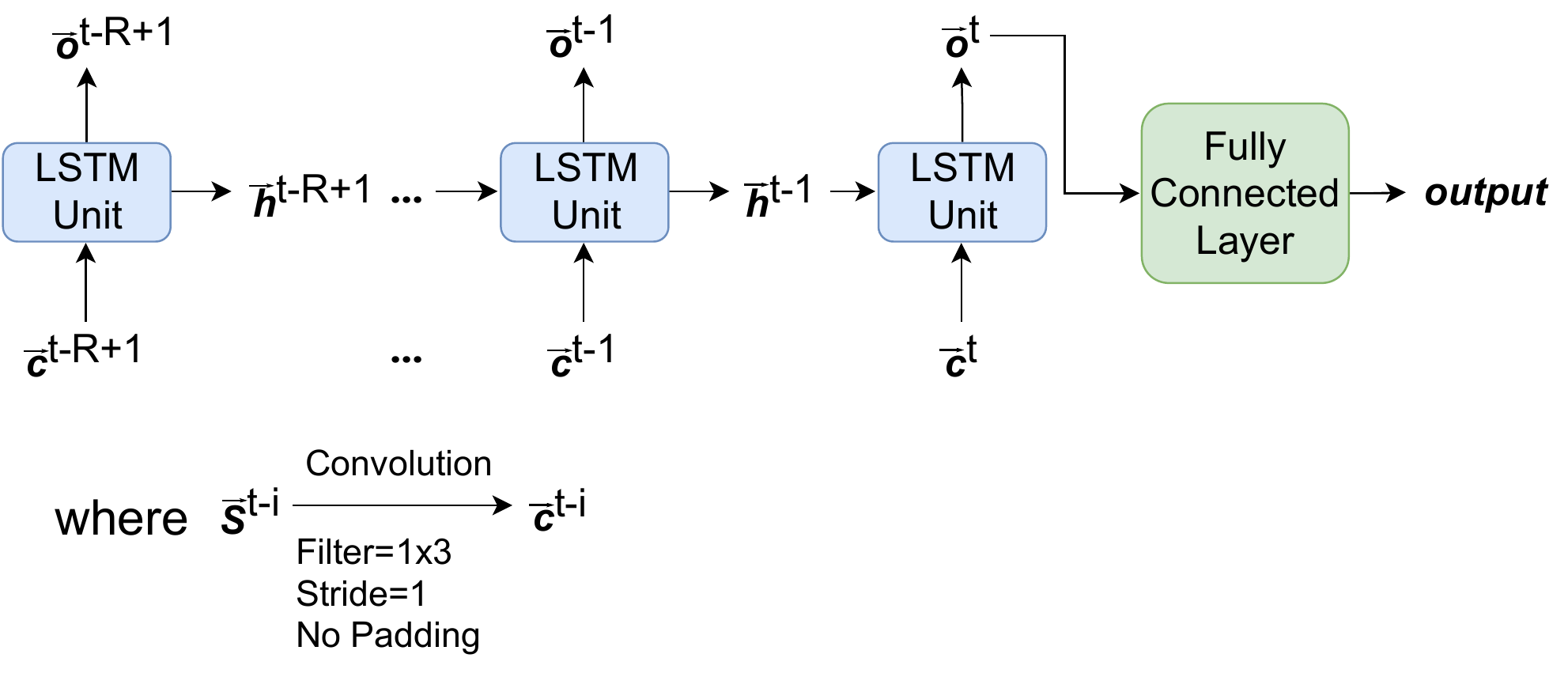}
	\caption{CNN-LSTM hybrid model for traffic flow prediction. }
	\label{Fig4_CNN_LSTM}
	\vspace{-.1cm}
\end{figure}

%
\begin{figure*}[h]
	\centering
	\subfloat[]{
		\includegraphics[height=0.173\textheight]{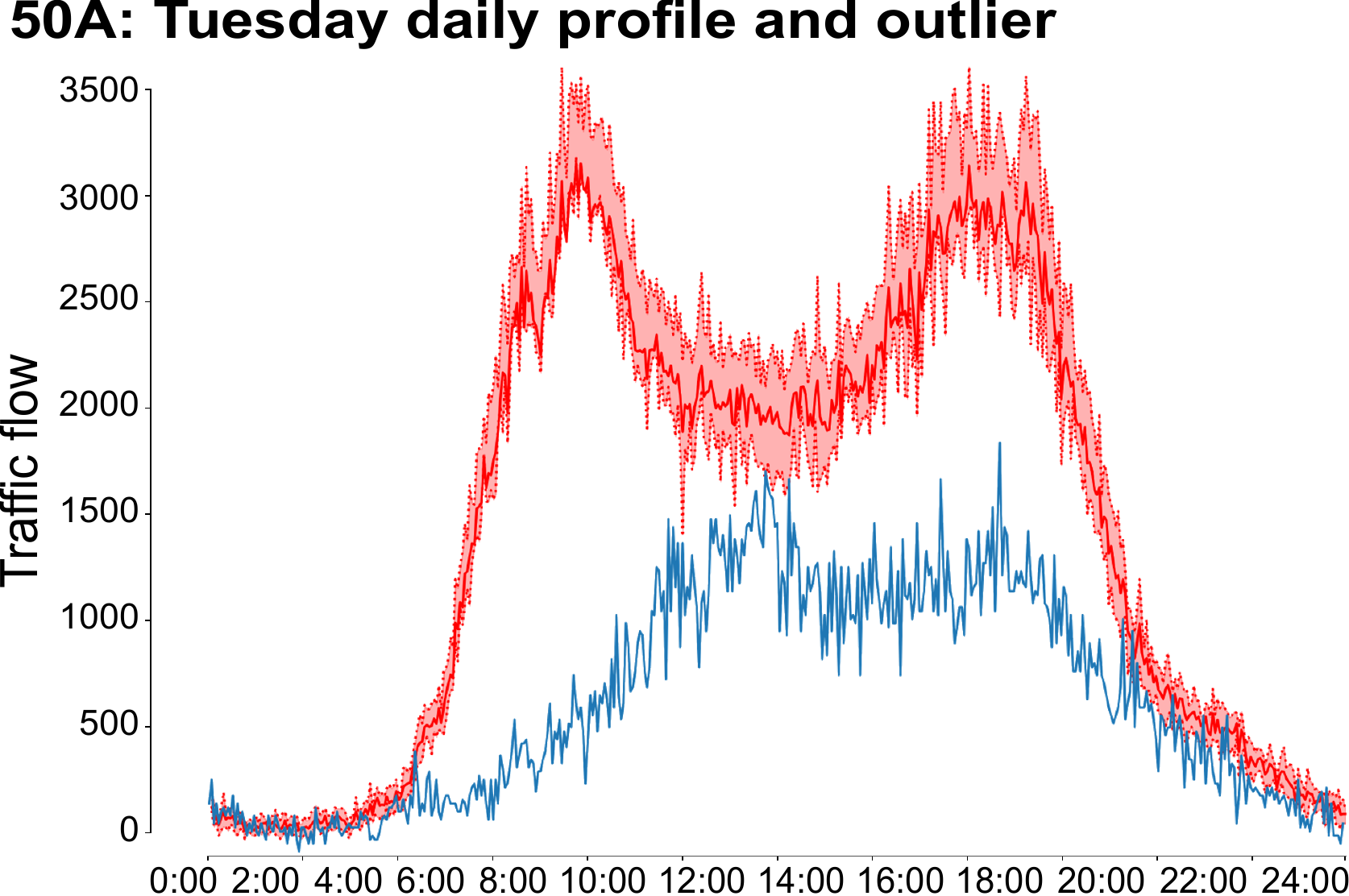}%
		\label{subfig:construct-daily-profile}%
	}%
		\subfloat[]{
		\includegraphics[height=0.173\textheight]{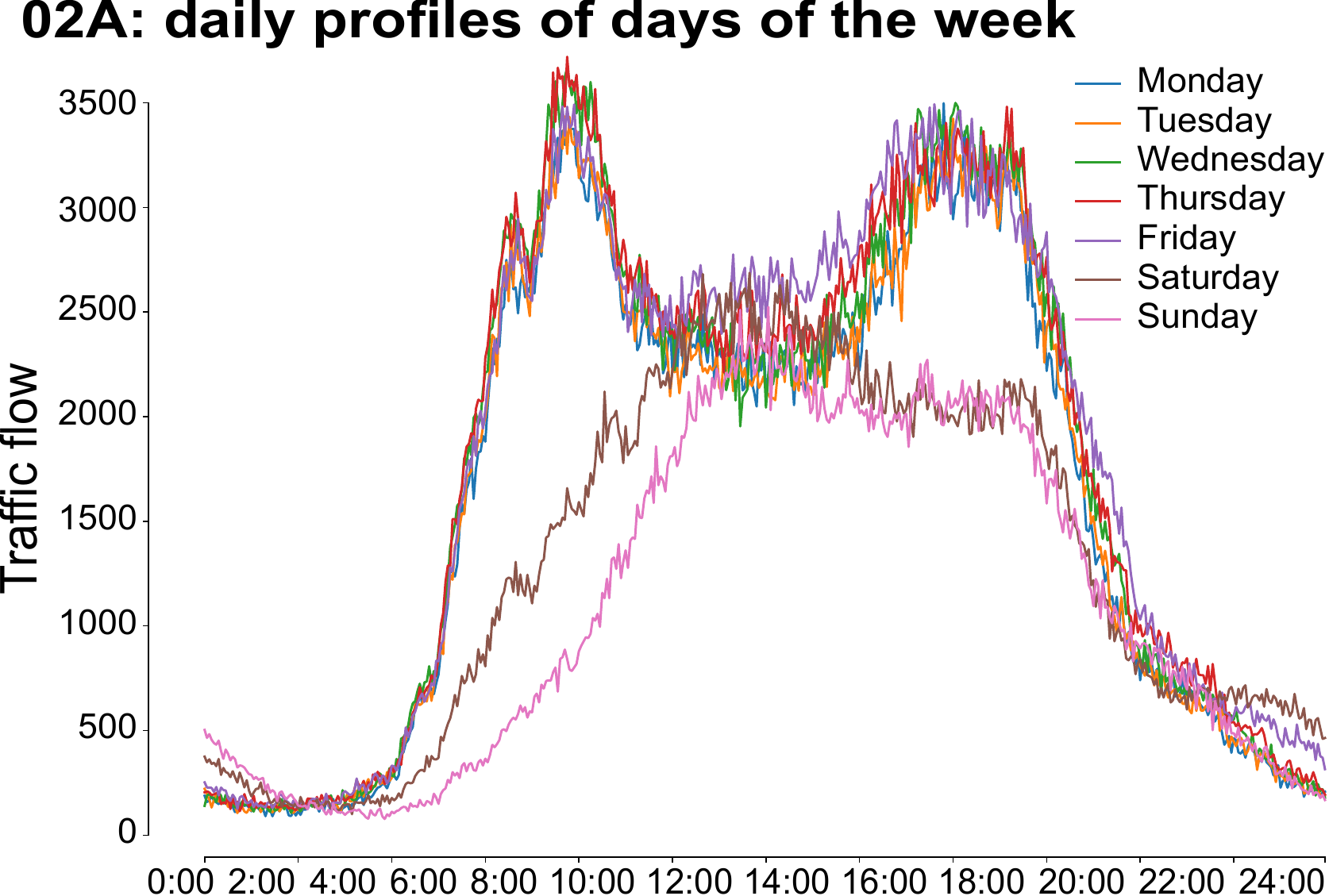}%
		\label{subfig:daily-profile-days-of-week}%
	}%
	\subfloat[]{
		\includegraphics[height=0.173\textheight]{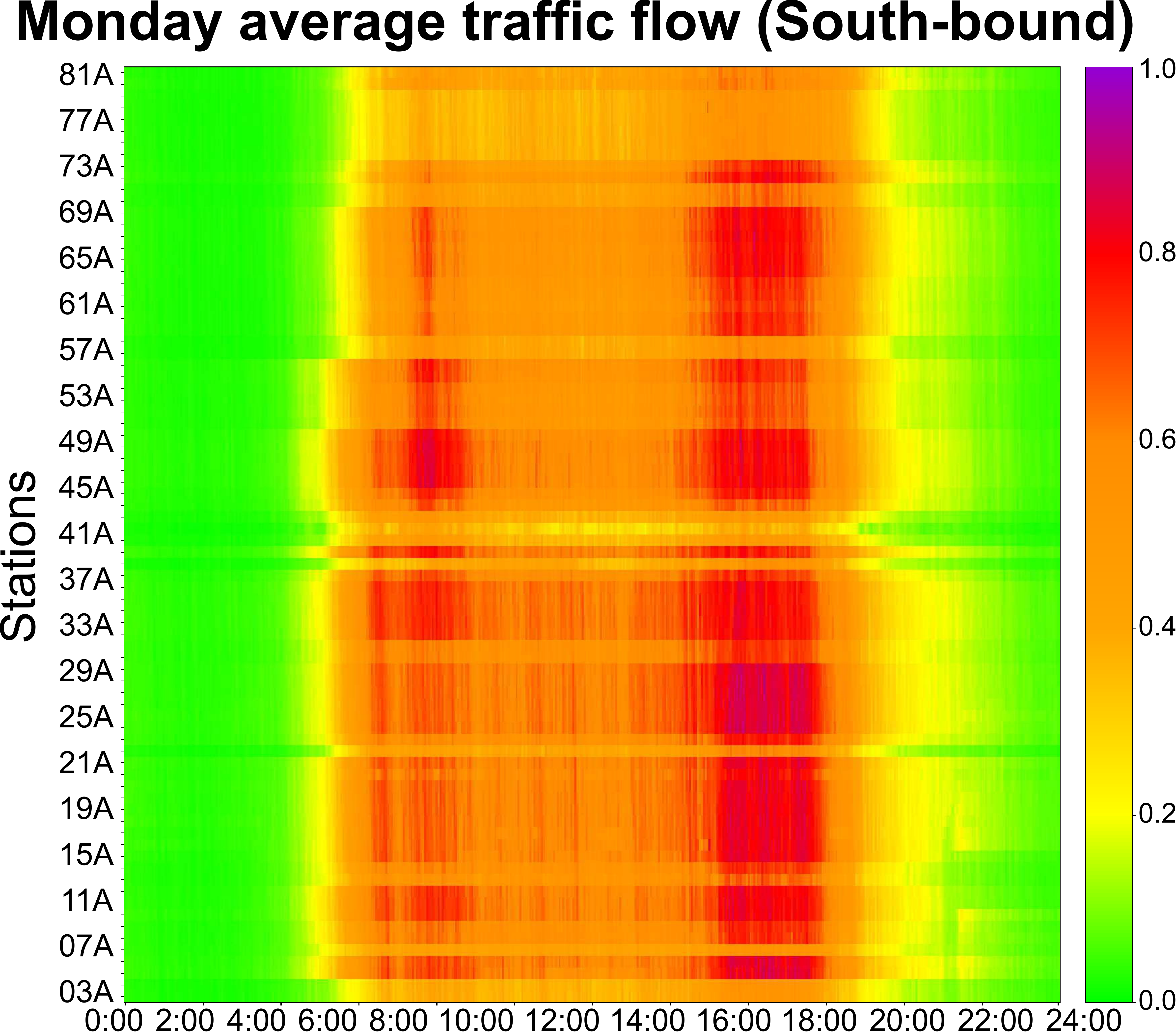}%
		\label{subfig:average-traffic-flow}%
	}%
	
	\caption{
      	\textbf{(a) Constructing the daily profile.} 
      	Mean (solid line) and the $20\%-80\%$ percentiles (red area) for the traffic flow series for the station \texttt{50A}, computed on the period 2017-02-01 to 04-30;
      	\textbf{(b) Daily profiles for days of the week.} 
      	The daily profiles for station \texttt{02A} for each of the days of the week, computed for the same period of time.
      	\textbf{(c) Daily profiles for all stations -- the Traffic Flow Congestion Map.} 
      	The colormap of the Monday traffic flow for all 104 south-bound (A) stations is calculated based on a Flow/Capacity ratio and ranges between 0 and 1.
	}
	\vspace{-.3cm}
\end{figure*}

\cref{Fig4_CNN_LSTM} showcases the structure of our hybrid model; the difference from a regular LSTM is that we use a 1-dimensional filter to scan the input (each $\boldsymbol{\vec{S}^{t-i}}$ is  processed by a 1-D convolution layer before it is fed into an LSTM unit). 
The added convolution layer has a 1x3 filter.

The results of each proposed model have been compared to other basic or parametric prediction models further detailed in \cref{S4_Exp}.

%
\section{Case study}\label{3_Case_study}

\subsection{The Sydney motorway traffic flow dataset}
\label{subsec:dataset}

Our methodology has been applied to a motorway traffic flow dataset, which was collected over the entire year of 2017, by recording the traffic flow at each of the 208 bi-directional ``flow counting stations'' along the M7 Motorway in Sydney, Australia (shown in \cref{Fig_0_Methodology}). 
The M7 motorway runs on the West of Sydney and it is the main motorway connecting North and South Sydney.
There are 104 metering stations in each direction including entries and exits; stations ending in A denote south-bound traffic, stations ending in B denote north-bound traffic while stations ending in E and X denote entries and exits respectively. The dataset contains 36.34 million data points, where one data point is the flow recorded by one station during one time-interval of 3min, denoted as \ti.

%
%
%

\subsection{Daily profile and traffic flow map}\label{subsec:Sec3_2_Daily_Profile}


\textbf{Daily profile.}
We start from the observation that the traffic flow at any given station presents a strong daily and weekly seasonality, mainly driven by the users daily work commute patterns.
We define the \emph{daily profile} as the typical daily traffic flow series recorded at a given station.
We compute a station's daily profile as the average flow for each \ti, for a given day of the week, over a period of three months (12 weeks in the period of February $1^{st}$ to April $30^{th}$ 2017).
\cref{subfig:construct-daily-profile} shows the mean flow for a Tuesday for the station \texttt{50A}, alongside with the $20^{th}$ and $80^{th}$ percentile values.
We make several observations.
First, the $20/80$ confidence interval wraps closely the daily profile, indicating that the 12 series are very similar and that the daily profile is a representative summarization.
Second, we observe that the daily profile shows two peaks, corresponding to the two rush hours: one in the morning (8-10 AM) corresponding to the daily commute towards work, and a second one in the afternoon (5-8 PM) corresponding to the end of the working day.
Last, the daily profile allows to identify non-standard days; for example, the blue line in \cref{subfig:construct-daily-profile} shows a significantly lower traffic, as it corresponds to April $25^{th}$ -- ANZAC day, a public holiday in Australia.

\textbf{Weekdays vs. weekend.}
\cref{subfig:daily-profile-days-of-week} plots the daily profiles for each of the days of the week, for station \texttt{02A}.
We observe that the weekdays (Monday to Friday) exhibit similar two peak patterns, whereas Saturday and Sunday have a single peak between 11AM and 4PM, and an overall lower flow.
Noteworthy, the ANZAC day flow shown earlier in \cref{subfig:construct-daily-profile} resembles a weekend pattern, despite being on a Tuesday.

\textbf{Traffic flow congestion map.}
\cref{subfig:average-traffic-flow} plots as a congestion map the Monday daily profiles of all South-bound stations by calculating the flow/capacity ratios. Here we also observe the two peak patterns of a typical weekday.
However, the traffic flow also allows to visually identify the most congested sections of the motorway (between \texttt{31A} and \texttt{13A} during the afternoon peak) and most importantly to track down abnormal congestion disruptions along the motorways. In addition to providing an abstraction of the typical flow, the daily profile is also used to correct the missing data in the dataset, most often occurring due to malfunctioning traffic recording devices for particular stations (see the online supplement~\cite{appendix} for more details). 
\section{Experimental setup}\label{S4_Exp}

In this section we describe the set-up, the implementation of the DL models and the comparison with other state-of-art prediction models. 

\paragraph{\textbf{Prediction setup}}
\label{D4_experiment_setup}

After outlier identification and missing point processing we select the time period of the DL models which comprises 8 months of the entire dataset (the other 4 months show an abnormally high volume of missing data and are, therefore, excluded from this analysis; see the online supplement~\cite[Fig. 9b]{appendix}). 
More specifically, we use the traffic flow from $01/02/17$ to $30/01/17$ and $01/06/17$ to $31/08/17$ respectively, for training the models (6 months in total). The flow from $01/09/17$-$30/09/17$ is used for validation while the flow from $01/10/17$-$31/10/17$ is used as a test set. This was kept for consistency across all DL models. 	
	
%
%

%
\begin{figure*}[h!]
	\centering
	\includegraphics[width=0.99\textwidth]{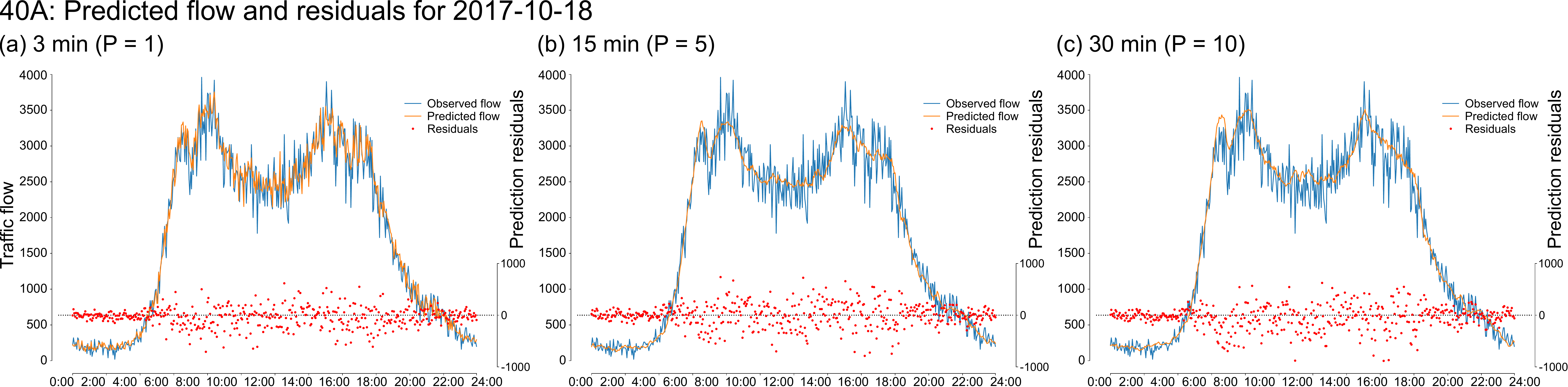}
	\caption{
		Observed and predicted traffic flow, and residuals for 3 min \textbf{(a)}, 15 min \textbf{(b)} and 30 min \textbf{(c)} for station \texttt{40A} on a weekday.}
	\label{Fig_6_LSTM_results}
\end{figure*}

\paragraph{\textbf{Other baseline models}}\label{D4_emodel_comparison} have been used for comparing the performance of selected DL models, such as:

\textbf{Daily Profile Prediction (DPP)}: is a base model in which we use the Daily Profile (described in \cref{subsec:Sec3_2_Daily_Profile}) computed for each station and each day of the week as a predictor; 
therefore for each station, we have 7 flow curves and each curve is consisting of 480 flow values (the time interval between counts is 3min). This model therefore predicts the average traffic flow per station. 

\textbf{BPNN for separate station prediction (Sep-BPNN)}: besides the BPNN model applied over all stations, we have also applied the prediction to individual stations as well, for showcasing how the prediction would be impacted if no information on neighbouring stations would be available to the models. Therefore this model is a combination of $N$-BPNN models consuming independently historical flow of each station, each using a hidden layer of 10 neurons.  
 
\textbf{ARIMA}: the ARIMA model predicts the next value in a series as a linear combination of the past observations.
We use a default ARIMA implementation, with the hyper-parameters $ARIMA(p=2,d=1,q=0)$.
Here,
$p$ is the parameter of the autoregression (note that $p$ in ARIMA has a different meaning than $P$ in \cref{tab:notations}), $d$ is for the degree of differencing (the number of times the data have had past values subtracted) and $q$ controls the moving average. 
We have determined the values of the hyper-parameters on the validation set, through exhaustive line search in the domains $p \in \{1, .. , 5\}$, $d \in \{1, .., 5\}$, and $q \in \{0, .., 3\}$.
We used 100 flow counts as the maximum historical time horizon for training. 
Note that ARIMA can only predict one value in the future.
To predict longer time horizons we add the prediction to the time series and we roll forward to the next \ti.
\paragraph{\textbf{DL implementation and hyper-parameter selection}}

\textbf{Past and future time horizon.}
At a given time point $t$, the input of each DL model is the traffic flow during the past $R$ time points, and the output is the prediction of the flow at the $P^{th}$ time point in the future.
Therefore, for any given station $j$ the input is $[x^{t-R+1}_j, x^{t-R+2}_j, \ldots, x^{t}_j]$, the output is $\overline{x^{t+P}_j}$ and the training performance is measured by how close the prediction is to the recorded flow $x^{t+P}_j$.
By varying $t$ on a dataset with $n$ time points, we obtain $n-R-P+1$ pairs of inputs and outputs.
Take for example $R=2$ and $P=1$ on a dataset with 5 time points.
The above procedure generates 3 train-test sets: [$x^1_j$,$x^2_j$]:$x^3_j$, [$x^2_j$,$x^3_j$]:$x^4_j$, [$x^3_j$,$x^4_j$]:$x^5_j$, where the column separates the training vector and the desired output. 
When $R=3$ and $P=2$, we have only one combination [$x^1_j$,$x^2_j$,$x^3_j$]:$x^5_j$ -- we predict the fifth time point based on the traffic flows during the first three data points.
Our dataset contains (42,721-R-P) + (44,161-R-P) combinations (as we have two separate contiguous training periods), the validation set contains (14,401-R-P) pairs and the test set contains (14,881-R-P) pairs.

%
%

\textbf{Avoiding overfitting.}
Even complex learners like our DL methods can overfit training data if trained for too long.
We control overfitting using the validation data-set.
At each DL epoch (i.e. learning iteration), we learn on the training set and we measure the performance on the validation set.
We record both the performance and the trained model after each epoch.
We terminate the training when the loss function on the validation dataset has not decreased for three consecutive epochs.
We select as the \emph{best trained model} (the stored model at the epoch that achieved the lowest validation error).
In practice, the training process ends in about 20-30 epochs.


\textbf{Deep learning parameters.}
We tune the values of the DL hyper-parameters on the validation set.
We vary the batch size in the range $[20,30,40,...75, 100]$ and we obtain a value of $50$. 
Our learning rate is $0.0003$ and the weight of the $L2$ regularisation term is $10^{-8}$.
All our DL models are implemented using PyTorch~\cite{paszke2017automatic}, using the Adam optimiser which provided a better performance than SGD or AdaGrad.
 
\paragraph{\textbf{Past and future prediction horizon selection}}
 
$R$ is an important hyper-parameter of our model -- the length of the learning past horizon -- which is tuned on the validation set in the range $R \in\{1,..,30\}$.
30 time points in the past corresponds to a 90 min past time horizon, which is more than the expected travel time along the whole M7 motorway in one direction.
Given a value of the prediction time horizon $P$, we train the model 5 times and we calculate the average accuracy on the validation dataset. 
\revA{We select as the \emph{best $R$}} the value that achieves the highest average accuracy for \revA{the} current $P$.
In \cref{R_and_P}, we focus on the relationship between $R$ and $P$ in order to answer several open questions: 
a) how much should we learn from the past to achieve best prediction results? 
b) how long in the future should we predict? 
c) is the size of the past horizon affecting the prediction results? 
d) what is the relation between $R$, $P$ \revA{and} the performances of the advanced DL models? 
%

\paragraph{\textbf{Model performance \revA{and training time}}}
We evaluate prediction performances using three widely used measures;
a) the Root Mean Square Error (RMSE), b) Mean Absolute Error (MAE) and c) Symmetric Mean Absolute Percentage Error (SMAPE). 
Due to space constraints, in \cref{subsec:performance} we only discuss RMSE results, the \revA{other convey the same results and they} are shown in the supplementary material~\cite{appendix}. 
\revA{All models are trained on an Intel Xeon processor with 24 cores, and they take between 10 and 15 epochs to converge.
For $P=5$ and $R=5$, the training time is $219.45\pm61.6$ (sec) for CNN, $302.10\pm100.9$ for LSTM and $382.53\pm107.7$ for CNN-LSTM; more information is provided in~\cite{appendix}.
}

%
\section{Main findings}\label{S5_Results}

In this section, we present the predictive comparative performance analysis for all models (\cref{subsec:performance}), and we analyse the interplay between the past ($R$) and future ($P$) time horizons (\cref{R_and_P}).

\subsection{Model performances}
\label{subsec:performance}

%
\begin{figure}[t]
	\centering
	\includegraphics[width=0.45\textwidth]{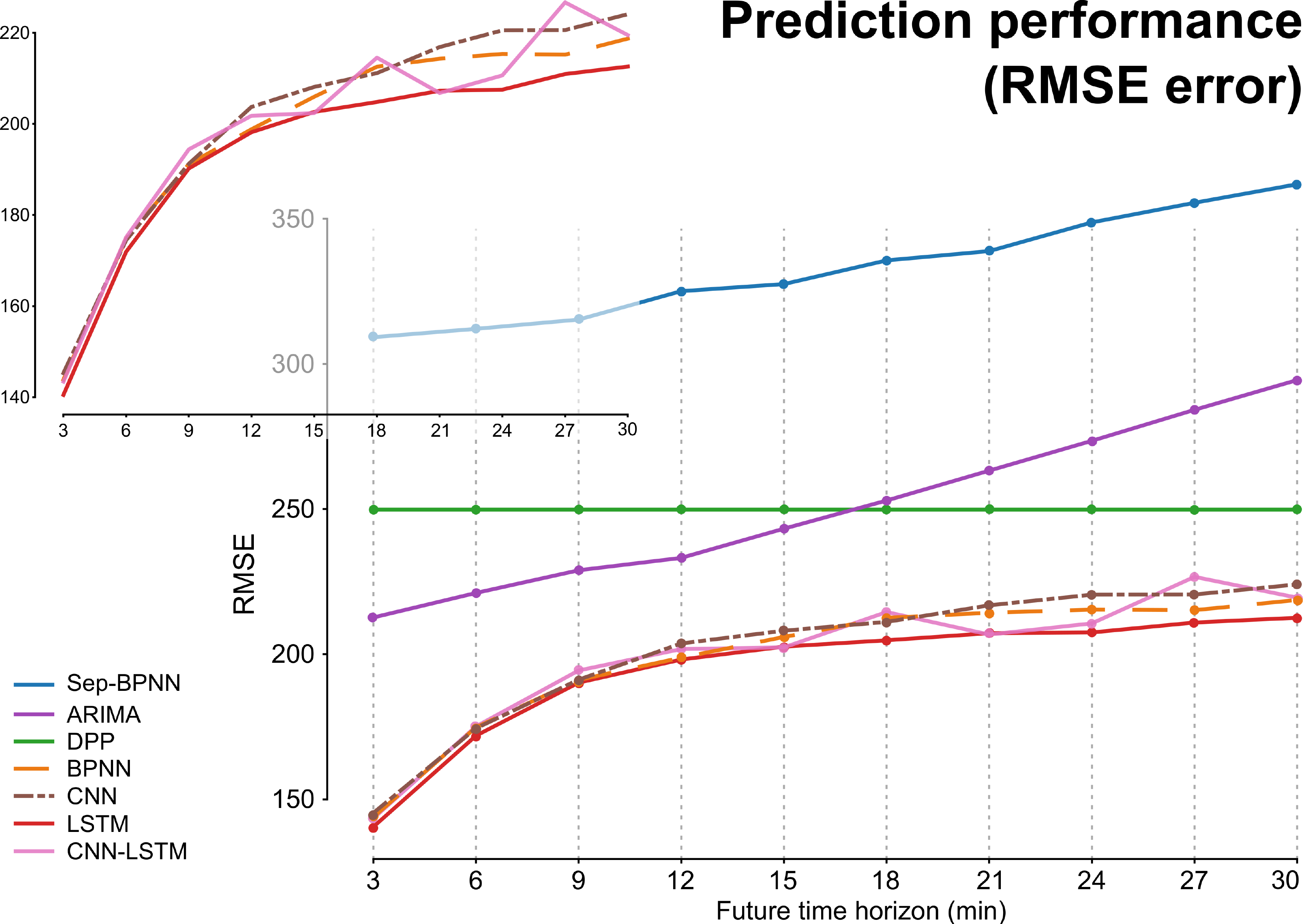}
	\caption{
		Prediction performance for all models (Oy axis), for increasing future time-horizons P (Ox axis). The zoom expands the performance of DL models. The y-axes show the RMSE of each model (lower is better).
	}
	\label{Fig_7_RMSE_all_models}
\end{figure}

\textbf{Prediction performance.}
We train all DL models and baselines on the testing and validation set, with varying prediction horizons $P \in \{1,..,10\}$, where $P=1$ represents a 3min prediction, while $P=10$ is equivalent of a 30min prediction in the future.  
\cref{Fig_7_RMSE_all_models} shows the RMSE prediction error for all models.
As expected, the prediction performance of all models (except DPP) decreases as we predict further into the future.
DPP (daily profile predictor) has a constant RMSE as it outputs historical averages for any data point.
Visibly, the worst performing model is the Sep-BPNN mostly because it does not incorporate the spatial and temporal correlation between the counting stations. 
The parametric model ARIMA appears to under-perform DPP for large prediction horizons, probably due to the accumulation of errors in its rolling prediction. 
The best performing models are the advanced DL models; 
LSTM outperforms all models for every $P$, followed closely by the hybrid model CNN-LSTM (which only for p=7 (21min) outperforms LSTM). 
The performances of the hybrid model fluctuate, and it is outperformed by regular CNN for $\{p\leqslant3$, $p=6$ and $p=9\}$ indicating that for our problem, a more sophisticated model does not necessarily improve performances. 
All DL models achieve similar performances for a prediction horizon lower than 12 min ($P \leq 4$), and performances stabilise after 21 min.

\textbf{Residuals analysis.}
\cref{Fig_6_LSTM_results} shows the observed and the predicted flow by LSTM (our best performer) during a day, for station \texttt{40A}.
We also show the prediction residuals (the difference between our prediction and the real flow count information). 
The prediction results are rolled out for the next 3min (\cref{Fig_6_LSTM_results}a), 15min (\cref{Fig_6_LSTM_results}b), and 30min into the future (\cref{Fig_6_LSTM_results}c); they show good performance for short-term predictions (less than 15min) with very low residuals outside peak hours and reaching a maximum error of $10.8\%$ during AM/PM peak intervals. 
For long-term prediction, the LSTM model maintains a good prediction performance for the overall traffic flow trend, but has lower accuracies for predicting smaller traffic flow deviations from main flow profile. 
The performance of all models has been investigated as well for flow prediction during stochastic events with major disruptions on the traffic network and it will be further presented in an extended version of this work. 

\subsection{Interplay between past and future horizons}\label{R_and_P}

%
\begin{figure*}[h!]
	\centering
	\includegraphics[width=0.99\textwidth]{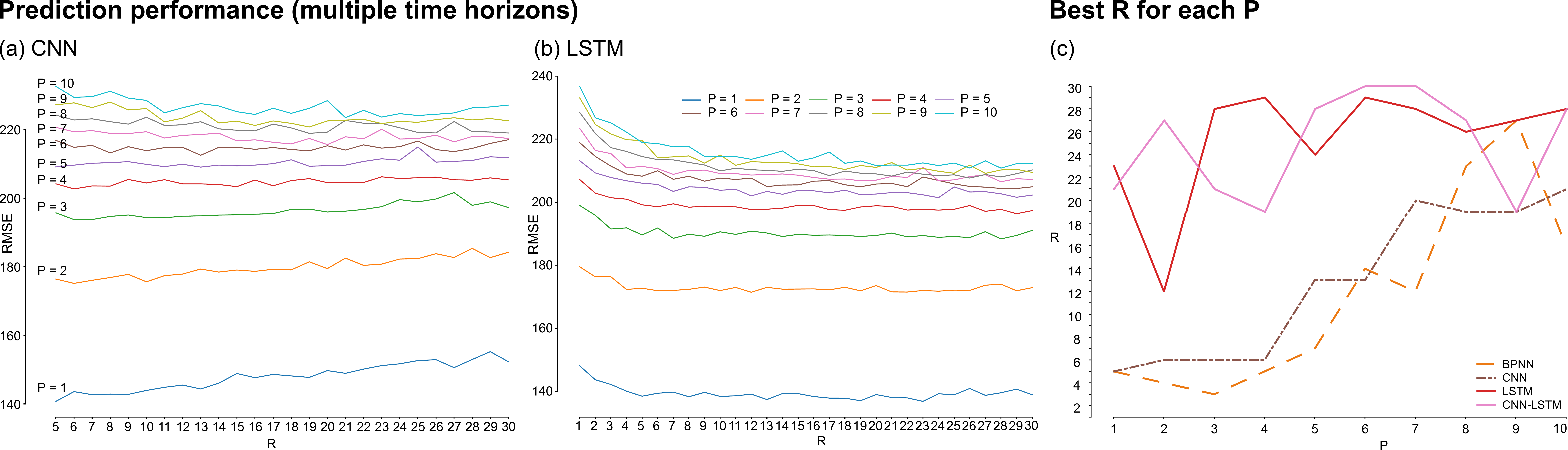}
	\caption{
		Prediction performances for multiple future time horizons for CNN \textbf{(a)} and LSTM \textbf{(b)} (lower is better).
		\textbf{(c)} Best length of past time horizon ($R$) for given future time horizons ($P$).
	}
	\label{Fig13_14_15}
\end{figure*}

%
%

\textbf{Predictions at multiple future time horizons:}
After evaluating performances, we next investigate the optimal past time horizon required by each DL model to make prediction at a given future point.
The past horizon extends up to 90min in the past ($R \leq 30$), while the explored future horizon reaches 30min in the future ($P \leq 10$).
\cref{Fig13_14_15}a and \cref{Fig13_14_15}b showcase the prediction performance over multiple time horizons for CNN and LSTM, the two most powerful DL models due to their capability to efficiently incorporate spatial and temporal features. 
Both models obtain their best RMSE for short time predictions ($P=1$), however CNN suffers a significant decrease in performance if we consider longer past horizons ($R>13=39min$).
This can be explained by the fact that CNN is designed for leveraging spatial correlations, and make less usage of temporal information. 
LTSM's performance improve with the availability of longer historical, admittedly slower for larger values of $R$.
Both models present a decreasing performance when we predict too far into the future ($P>5=15min$), and RMSE appears not to decrease significantly for large values of $P$.
CNN-LSTM behaves very similarly to LSTM (shown in the online supplement~\cite{appendix}), indicating the the hybrid model predictions are dominated largely by the temporal component.

\textbf{Best $R$ for $P$:}
To complete our analysis, in \cref{Fig13_14_15}c we showcase the best past horizon dimension which is selected for each future prediction horizon, across each DL model.
We observe that LSTM and the hybrid CNN-LSTM make use of larger past time horizons even when making short-term predictions.
When predicting 9min ahead ($P=3$), the best LSTM performance leverages 69min in the past ($R=23)$, whereas CNN only uses 18min in the past ($R=6$).
This results again reinforces the fact that LSTM and CNN-LSTM can learn long-term trends to make more accurate predictions.
Though not our case, it may prove problematic when long historical data is not available, in which case CNN and BPNN might provide better results.
\section{CONCLUSIONS}\label{S6_Conclusions}
This paper presents an advanced DL framework for motorway traffic flow prediction,  by chaining together data profiling and outlier identification, spatial and temporal feature generation and various DL model development.
The current approach has been applied on 36.34 million data points, to make traffic flow prediction over an entire motorway in Sydney Australia. 
The findings showcase LSTM as having the best predictive performance, despite having competed against a hybrid model combining CNN  and LSTM.
Our analysis reveals that the optimal past time horizon needs to be adapted for each DL model: LSTM and its variants learn long-term trends and require longer histories, while CNN learns spatial correlations from short histories.
\verify{Starting from the initial 3 challenges listed in \cref{S1_Intro}, we summarise the advantages of our proposed deep learning modelling: 
a) it provides good prediction accuracy for a large number of counting stations, 
b) its usage is based on a tailored selection of past learning horizon and future prediction horizon and 
c) the hybrid CNN-LSTM model under-performed when compared to individual LSTM, which indicates that the more complex deep learning models do not improve the prediction accuracy for our motorway flow prediction study.}

Our \verify{future} work includes designing traffic flow-based detection methods for
stochastic events which can massively disrupt the traffic flow along motorways.
For larger transport networks, encompassing larger areas and complex structures, an alternative that we are exploring is to explicitly incorporate the spatial relations between traffic stations using CNNs with graph structured data (see \cite{YAN2019259}, \cite{Henaff2015DeepCN}).




{ \small
\section*{ACKNOWLEDGMENT}
This work has been done as part of the ARC Linkage Project LP180100114. 
The authors are highly grateful for the support of Transport for NSW, Australia. 
}


\bibliographystyle{IEEEtran}
\bibliography{IEEE_ITSC_2019}

\begin{thebibliography}{10}
\providecommand{\url}[1]{#1}
\csname url@rmstyle\endcsname
\providecommand{\newblock}{\relax}
\providecommand{\bibinfo}[2]{#2}
\providecommand\BIBentrySTDinterwordspacing{\spaceskip=0pt\relax}
\providecommand\BIBentryALTinterwordstretchfactor{4}
\providecommand\BIBentryALTinterwordspacing{\spaceskip=\fontdimen2\font plus
\BIBentryALTinterwordstretchfactor\fontdimen3\font minus
  \fontdimen4\font\relax}
\providecommand\BIBforeignlanguage[2]{{%
\expandafter\ifx\csname l@#1\endcsname\relax
\typeout{** WARNING: IEEEtran.bst: No hyphenation pattern has been}%
\typeout{** loaded for the language `#1'. Using the pattern for}%
\typeout{** the default language instead.}%
\else
\language=\csname l@#1\endcsname
\fi
#2}}

\bibitem{appendix}
online supplement, ``Appendix: \titlename,'' 2019,
  \url{https://arxiv.org/pdf/1907.06356.pdf#page=9}.

\bibitem{VANDERVOORT1996307}
M.~V.~D. Voort, M.~Dougherty, and S.~Watson, ``Combining kohonen maps with
  arima time series models to forecast traffic flow,'' \emph{Transportation
  Research Part C}, vol.~4, no.~5, pp. 307 -- 318, 1996.

\bibitem{dKamarianakis2003}
Y.~Kamarianakis and P.~Prastacos, ``Forecasting traffic flow conditions in an
  urban network: Comparison of multivariate and univariate approaches,''
  \emph{Trans. Res. Rec.}, vol. 1857, no.~1, pp. 74--84, 2003.

\bibitem{quteprints63176}
A.~M. Khoei, A.~Bhaskar, and E.~Chung, ``Travel time prediction on signalised
  urban arterials by applying sarima modelling on bluetooth data,'' in
  \emph{Australasian Transport Research Forum}, 2013.

\bibitem{Williams2001}
B.~M. Williams, ``Multivariate vehicular traffic flow prediction: Evaluation of
  arimax modeling,'' \emph{Trans. Res. Rec.}, vol. 1776, no.~1, pp. 194--200,
  2001.

\bibitem{Chang2012}
H.~{Chang}, Y.~{Lee}, B.~{Yoon}, and S.~{Baek}, ``Dynamic near-term traffic
  flow prediction: system-oriented approach based on past experiences,''
  \emph{IET Intel. Transport Systems}, vol.~6, no.~3, pp. 292--305, 2012.

\bibitem{Jeong2013}
Y.~{Jeong}, Y.~{Byon}, M.~M. {Castro-Neto}, and S.~M. {Easa}, ``Supervised
  weighting-online learning algorithm for short-term traffic flow prediction,''
  \emph{IEEE Transactions on Intelligent Transportation Systems}, vol.~14,
  no.~4, pp. 1700--1707, 2013.

\bibitem{KARLAFTIS2011}
M.~Karlaftis and E.~Vlahogianni, ``Statistical methods versus neural networks
  in transportation research: Differences, similarities and some insights,''
  \emph{Trans. Research Part C}, vol.~19, no.~3, pp. 387 -- 399, 2011.

\bibitem{IDe2009}
D.~{Ide} and S.~{Kato}, ``Travel-time prediction using gaussian process
  regression: A trajectory,'' in \emph{International Conference on Data
  Mining}, 2009, pp. 1185--1196.

\bibitem{Ali2018}
U.~Ali and T.~Mahmood, ``Using deep learning to predict short term traffic
  flow: A systematic literature review,'' \emph{Intelligent Transport Systems},
  pp. 90 -- 101, 2018.

\bibitem{Nguyen2018}
H.~Nguyen, M.~Kieu, T.~Wen, and C.~Cai, ``Deep learning methods in
  transportation domain: A review,'' \emph{IET Intelligent Transport Systems},
  vol.~12, 07 2018.

\bibitem{TIAN2018297}
Y.~Tian, K.~Zhang, J.~Li, X.~Lin, and B.~Yang, ``Lstm-based traffic flow
  prediction with missing data,'' \emph{Neurocomputing}, vol. 318, pp. 297 --
  305, 2018.

\bibitem{Nguyen2019}
H.~Nguyen, C.~Bentley, M.~Kieu, Y.~Fu, and C.~Cai, ``A deep learning system for
  travel speed predictions on multiple arterial road segments,'' 02 2019.

\bibitem{Wu2019}
Y.~Wu, H.~Tan, L.~Qin, B.~Ran, and Z.~Jiang, ``A hybrid deep learning based
  traffic flow prediction method and its understanding,'' \emph{Transportation
  Research Part C}, vol.~90, p. 166–180, 2018.

\bibitem{ZHANG201465}
Y.~Zhang, Y.~Zhang, and A.~Haghani, ``A hybrid short-term traffic flow
  forecasting method based on spectral analysis and statistical volatility
  model,'' \emph{Transportation Research Part C}, vol.~43, pp. 65 -- 78, 2014.

\bibitem{tensorflow2015}
M.~Abadi, A.~Agarwal, P.~Barham, E.~Brevdo, Z.~Chen, C.~Citro, G.~S. Corrado,
  A.~Davis, J.~Dean, M.~Devin, S.~Ghemawat, \emph{et~al.}, ``{TensorFlow}:
  Large-scale machine learning on heterogeneous systems,'' 2015.

\bibitem{paszke2017automatic}
A.~Paszke, S.~Gross, S.~Chintala, G.~Chanan, E.~Yang, Z.~DeVito, Z.~Lin,
  A.~Desmaison, L.~Antiga, and A.~Lerer, ``Automatic differentiation in
  pytorch,'' in \emph{NIPS-W}, 2017.

\bibitem{Tan2018}
H.~Tan, L.~Qin, Z.~Jiang, Y.~Wu, and B.~Ran, ``{A hybrid deep learning based
  traffic flow prediction method and its understanding},'' \emph{Trans.
  Research Part C}, vol.~90, no. March, pp. 166--180, 2018.

\bibitem{POLSON20171}
N.~G. Polson and V.~O. Sokolov, ``Deep learning for short-term traffic flow
  prediction,'' \emph{Trans. Research Part C}, vol.~79, pp. 1 -- 17, 2017.

\bibitem{Wang2016TrafficSP}
J.~Wang, Q.~Gu, J.~Wu, G.~Liu, and Z.~Xiong, ``Traffic speed prediction and
  congestion source exploration: A deep learning method,'' \emph{International
  Conference on Data Mining}, pp. 499--508, 2016.

\bibitem{CHENG2010482}
X.~Cheng, W.~Lin, E.~Liu, and D.~Gu, ``Highway traffic incident detection based
  on bpnn,'' \emph{Procedia Engineering}, vol.~7, pp. 482 -- 489, 2010, 2010
  Symp. on Security Detection and Information Processing.

\bibitem{LI2018116}
S.~Li, T.~Chen, L.~Wang, and C.~Ming, ``Effective tourist volume forecasting
  supported by pca and improved bpnn using baidu index,'' \emph{Tourism
  Management}, vol.~68, pp. 116 -- 126, 2018.

\bibitem{AAAI1714501}
J.~Zhang, Y.~Zheng, and D.~Qi, ``Deep spatio-temporal residual networks for
  citywide crowd flows prediction,'' in \emph{AAAI Conference on Artificial
  Intelligence}, 2017.

\bibitem{NIPS2012_4824}
A.~Krizhevsky, I.~Sutskever, and G.~E. Hinton, ``Imagenet classification with
  deep convolutional neural networks,'' in \emph{Advances in Neural Information
  Processing Systems 25}, 2012, pp. 1097--1105.

\bibitem{Kang2017}
D.~{Kang}, Y.~{Lv}, and Y.~{Chen}, ``Short-term traffic flow prediction with
  lstm recurrent neural network,'' in \emph{International Conference on
  Intelligent Transportation Systems}, Oct 2017, pp. 1--6.

\bibitem{YANG2019320}
B.~Yang, S.~Sun, J.~Li, X.~Lin, and Y.~Tian, ``Traffic flow prediction using
  {LSTM} with feature enhancement,'' \emph{Neurocomputing}, vol. 332, pp. 320
  -- 327, 2019.

\bibitem{YAN2019259}
X.~Yan, T.~Ai, M.~Yang, and H.~Yin, ``A graph convolutional neural network for
  classification of building patterns using spatial vector data,'' \emph{ISPRS
  Journal of Photogrammetry and Remote Sensing}, vol. 150, pp. 259 -- 273,
  2019.

\bibitem{Henaff2015DeepCN}
M.~Henaff, J.~Bruna, and Y.~LeCun, ``Deep convolutional networks on
  graph-structured data,'' \emph{CoRR}, vol. abs/1506.05163, 2015.

\end{thebibliography}

%
\clearpage
\appendix

This document is accompanying the submission \textit{\titlename}.
The information in this document complements the submission, and it is presented here for completeness reasons.
It is not required for understanding the main paper, nor for reproducing the results.
%

\section*{Missing traffic flow data and data correction}

\begin{figure}[h!]
	\centering
	\subfloat[]{%
		\includegraphics[scale=0.18]{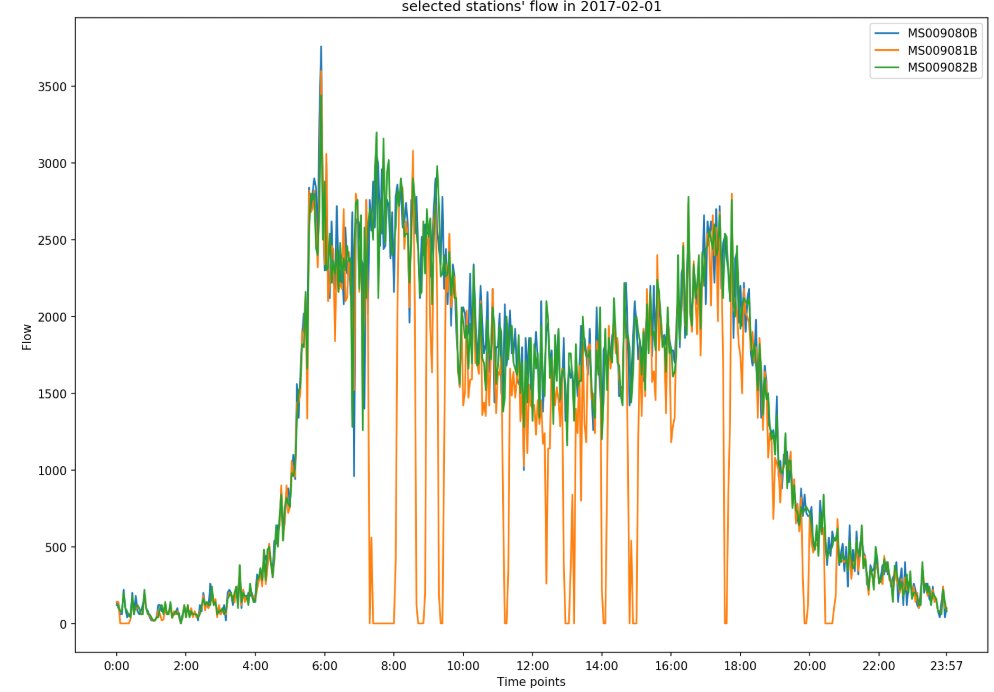}%
		\label{si-subfig:missing-data-example}%
	} \\
	\subfloat[]{%
		\includegraphics[scale=0.18]{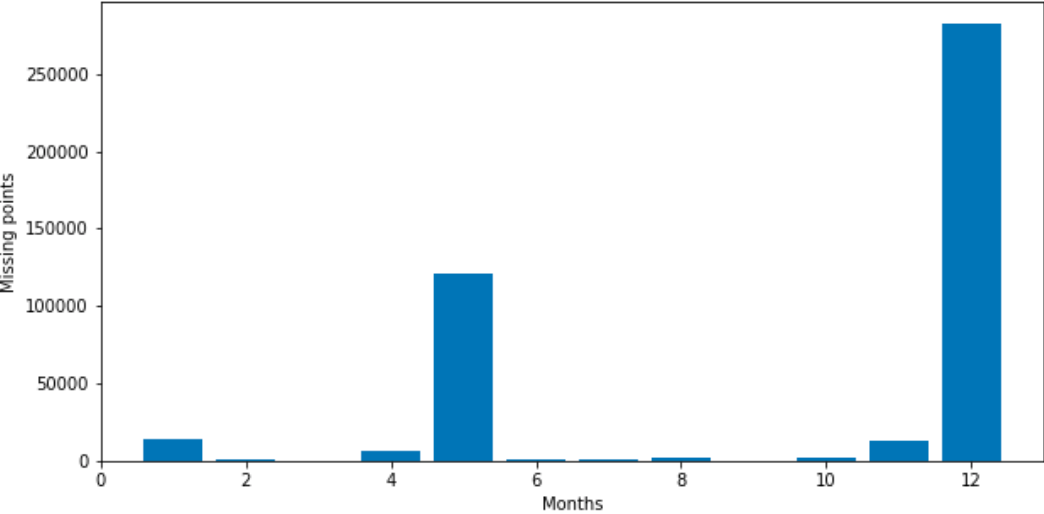}%
		\label{si-subfig:missing-data-counts}%
	}%
		\caption{
		Missing data in the Sydney Motorway traffic flow dataset.
		\textbf{(a)} The traffic flow for 1$^{st}$ of February 2017, for three contiguous stations (\texttt{80B}, \texttt{81B} and \texttt{82B}) with no entries and exits in between.
		\texttt{81B} is showing missing data.
		\textbf{(b)} The total number of missing data points, aggregated per month.}
\end{figure}

Upon inspection, the Sydney Motorway traffic flow dataset contains missing data.
The missing data is recorded as a traffic flow of zero, despite it clearly being incorrect.
\cref{si-subfig:missing-data-example} shows an example of such missing data.
\texttt{80B}, \texttt{81B} and \texttt{82B} are three stations, following contiguously one after the other.
We note that these three stations are of the type \emph{Passing traffic} in \cref{Fig_0_Methodology}b, i.e. there are no entries and no exits in between them.
It should follow logically that the traffic on all three stations should be similar.
However, \cref{si-subfig:missing-data-example} shows sudden drops of counts for station \texttt{81B} -- the station in between the other two.
The logical consequence is that the recording mechanism of station \texttt{81B} has malfunctioned during those periods, leading to recording a traffic of zero.
%
%

We detect such malfunctions at the level of the entire dataset, and we count them per month.
\cref{si-subfig:missing-data-counts} shows that there is an abnormally high number of missing data in the months of May and December 2017.
We therefore exclude these two months from our training data (as discussed in \cref{S4_Exp} of the main paper).


We correct missing data such as shown in \cref{si-subfig:missing-data-example} by interpolation.
We implemented and tested several approaches.
The first approach is to copy the traffic flow value from the same \ti of the previous or next day. 
However, as shown in \cref{subfig:daily-profile-days-of-week}, there are significant differences between the different days of the week, and especially between weekdays and weekends.
The second approach is to compute the average of the values at the previous and subsequent \ti (or a window around the missing data point).
However, we find cases with multiple contiguous missing data points (sometimes more than 10 in a row), probably due to extended malfunctions of the reporting equipment.
This renders the interpolation inaccurate.
\\
The third option that we considered is using 
the average flow at the same \ti from the same day of all weeks in a month.
For example, for correcting a missing data point at 2:00 pm in April 3rd (which is a Monday), we compute the average of the data points at 2:00 pm on all Mondays of April (Apr 10th, 17th, 24th) and use this value to fill the missing value.
This is equivalent to using the daily profile computed on the month in question.
We find the third approach to perform best and we use it throughout all experiments.

\begin{table}[tbp]
	\caption{The time spent on training our models [sec]}
	\centering
	\begin{tabular}{cccccc}
		\toprule
		& BPNN& CNN& LSTM& CNN-LSTM\\
		\midrule
		Mean & 101.190 &219.452 &302.105 &382.538\\ 
		Std &28.304 &61.610 &100.960 &107.722\\
		\bottomrule
	\end{tabular}
	\label{tab:execution-time}
\end{table}

\section*{Training time analysis}
\begin{figure}[htbp]
	\centering
	\subfloat[]{%
		\includegraphics[width=0.45\textwidth]{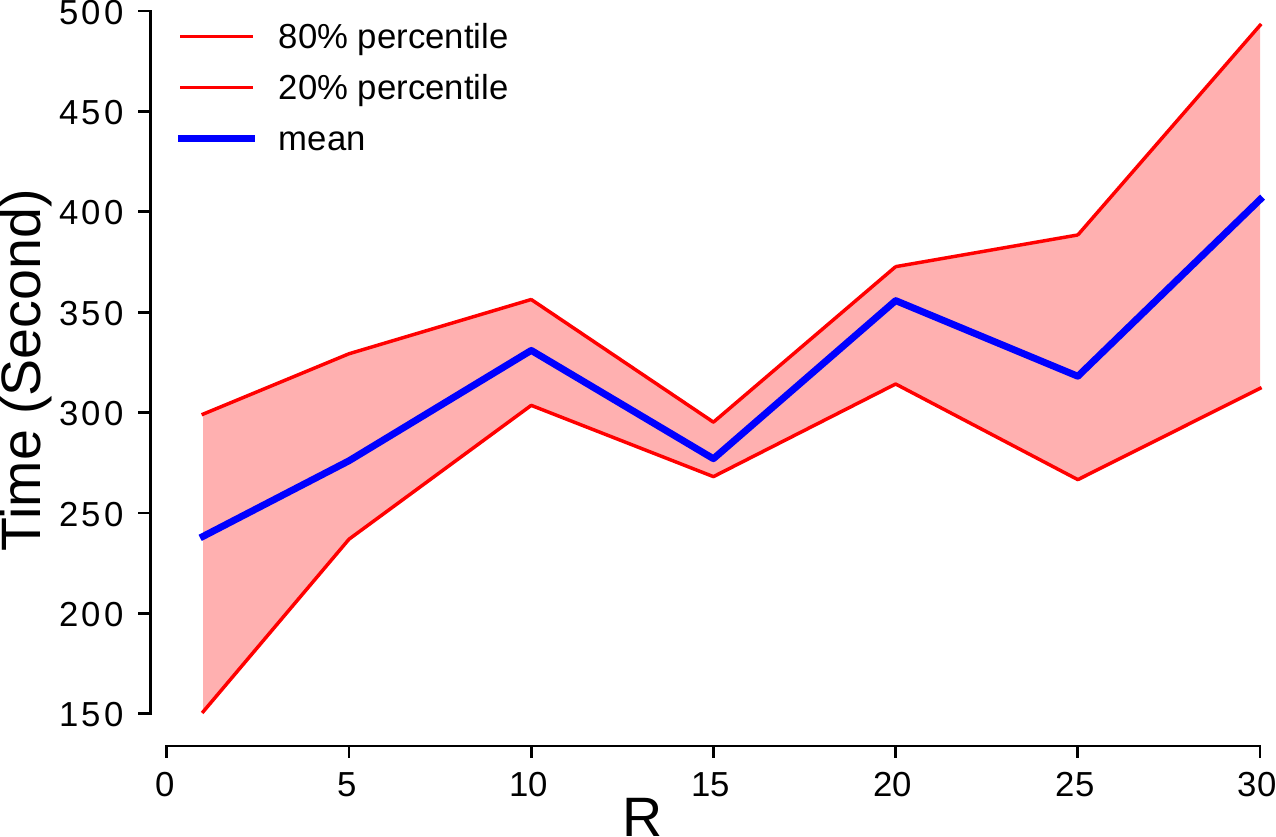}%
		\label{si-subfig:lstm-time}%
	} \\%
	\subfloat[]{%
		\includegraphics[width=0.45\textwidth]{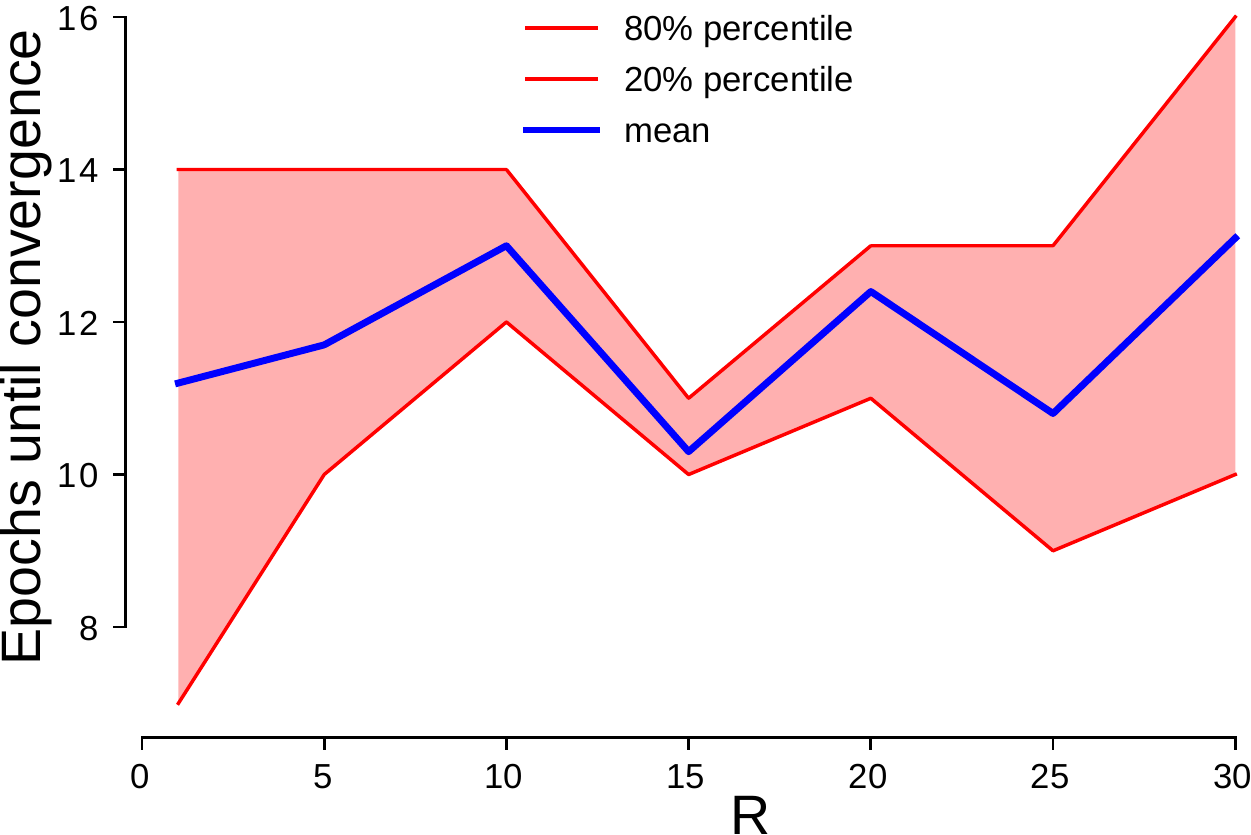}%
		\label{si-subfig:lstm-epoch}%
	}%
	\caption{
		Training time \textbf{(a)} and epochs to convergence \textbf{(b)} required by LSTM, with multiple values of $R$.
		The shaded area indicates the $20\%-80\%$ percentiles interval.
	}
\end{figure}

\revA{In this section, we evaluate the time needed to train our advanced DL models.
The main factors that influence the training time are $R$ -- controlling the extent of the past horizon taken into account when predicting the future -- and the number of epochs required for the DL methods to converge.
$R$ is a hyperparameter of the model, larger values of $R$ require more training time per epoch.
The number of epochs until convergence are data dependent.
\cref{si-subfig:lstm-time,si-subfig:lstm-epoch} illustrate for the LSTM model the relation between the value of $R$ and the training time (\cref{si-subfig:lstm-time}), and the number of epochs until convergence and the required training time (\cref{si-subfig:lstm-epoch}). 
For each value of $R$, we repeat the training of the model 10 times, and we show the mean training time and the standard deviation.
Visibly, the training time typically increases with the value of R increase, except for $R = 15$.
We observe the mean number of epoch required to converge fluctuates between 11 and 13, which implies that the value of $R$ does not affect the number of required epochs until convergence. 
Interestingly enough, both the mean and the standard deviation of the number of epochs until convergence drops significantly for $R = 15$, which also explains the drop in training time.
The reason of this drop would require further investigation and it is part of our future work plan.
\cref{tab:execution-time} shows the mean and standard deviation for the training time for all four DL models, for $P=5$ and $R=5$.
Each training for each model is repeated 20 times.
We find that BPNN is the fastest to train, followed by CNN and LSTM.
The complex model CNN-LSTM is the slowest to train.
}

\section*{Additional graphics}

\begin{figure}[tb]
	\centering
	\subfloat[]{%
		\includegraphics[width=0.45\textwidth]{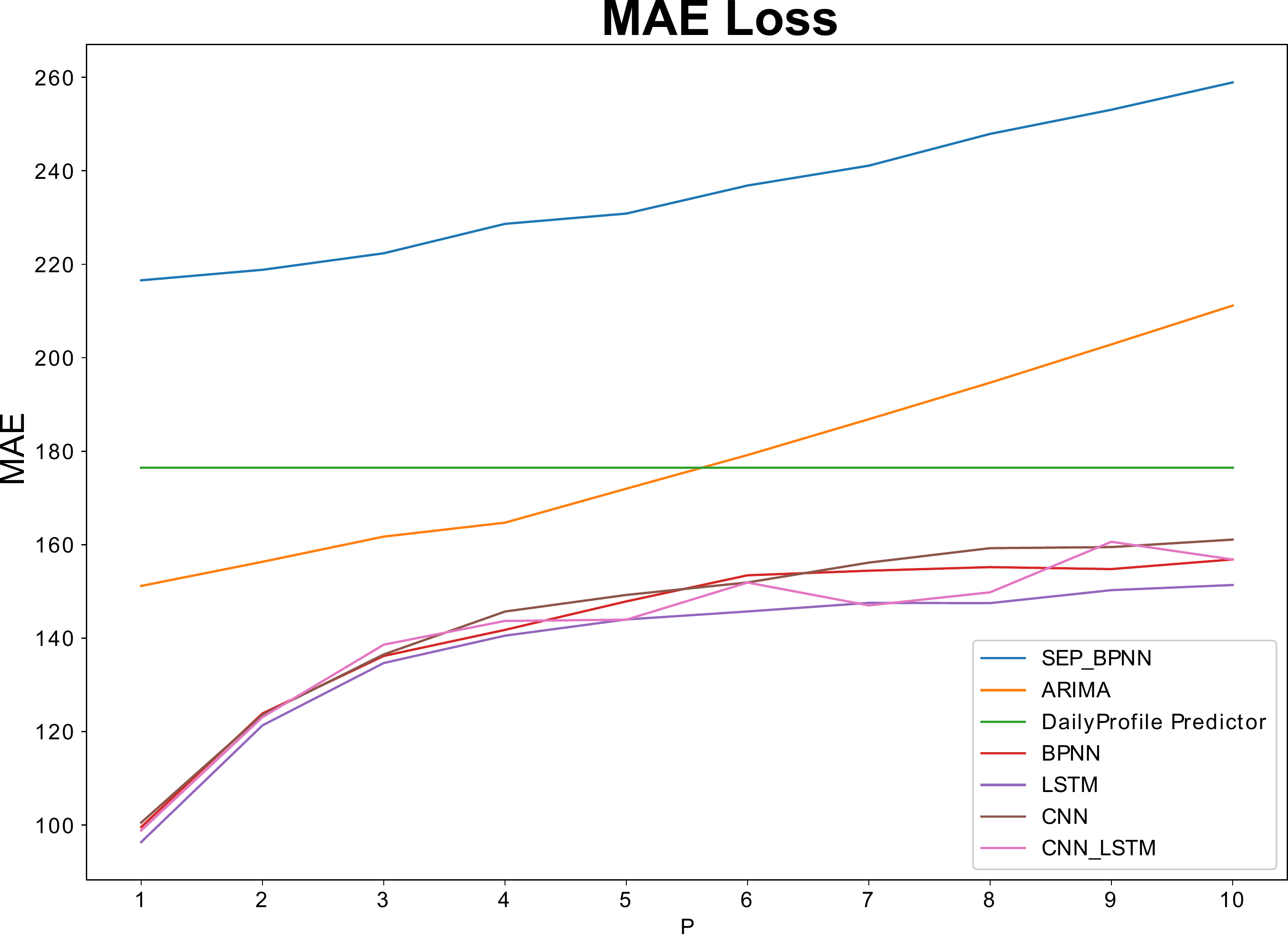}
		\label{si-subfig:mae-loss}%
	} \\%
	\subfloat[]{%
		\includegraphics[width=0.45\textwidth]{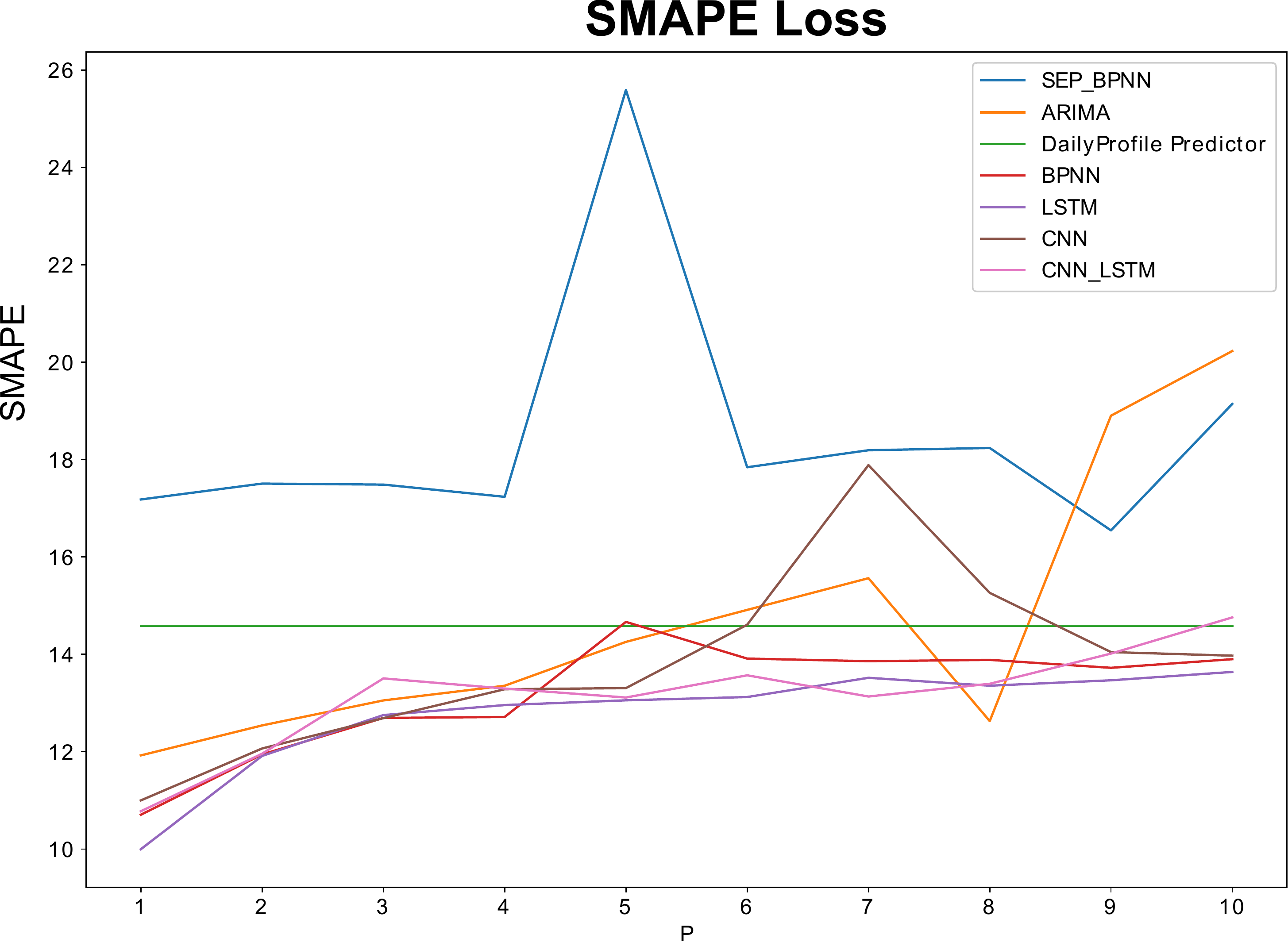}
		\label{si-subfig:smape-loss}%
	} \\%
	\subfloat[]{%
		\includegraphics[width=0.45\textwidth]{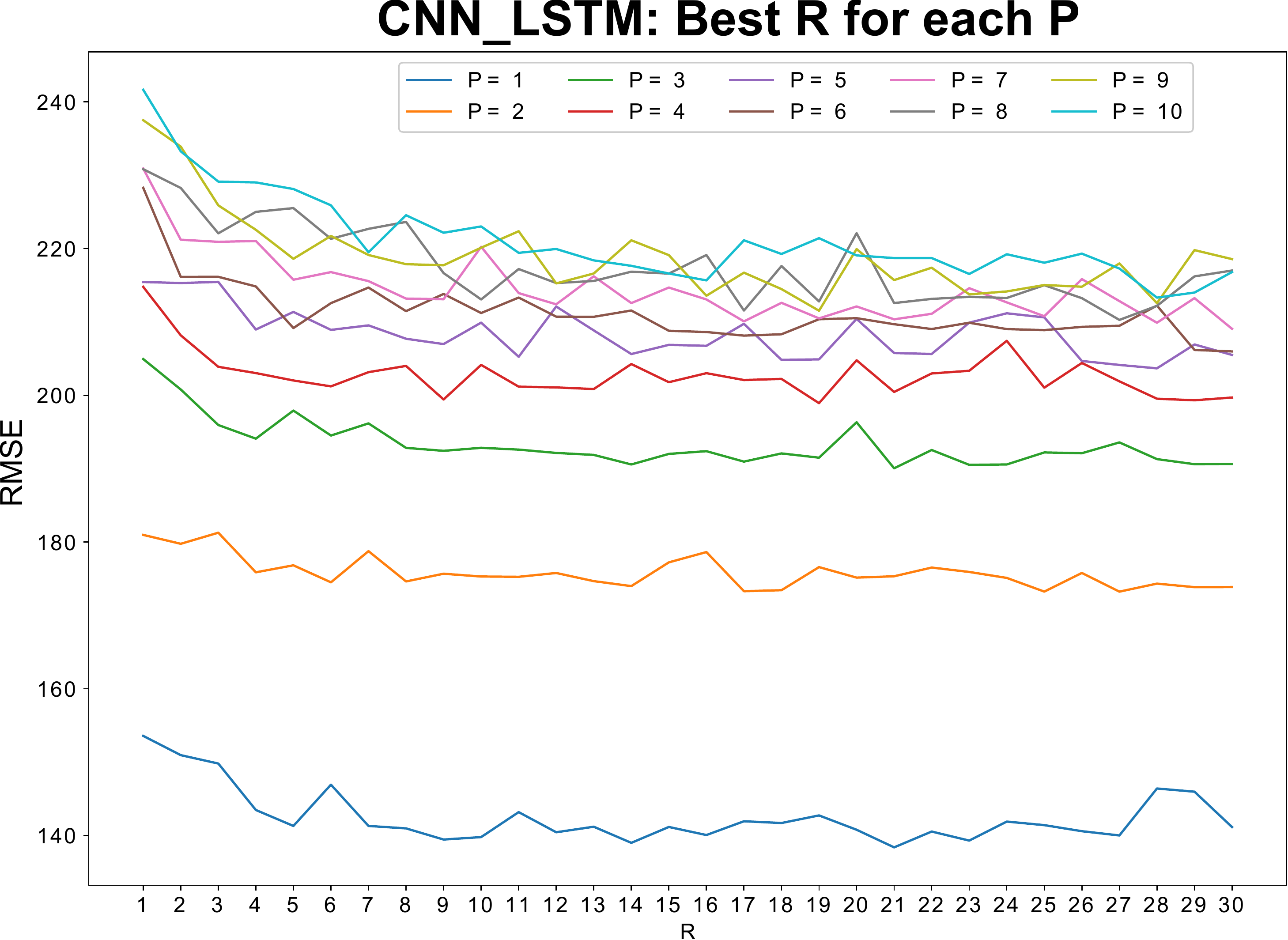}
		\label{si-subfig:CNN-LSTM-best-R-for-P}%
	}%
	\newline
	\caption{a) MAE loss results calculated across all models b) SMAPE error calculated for all comparative models present in this study and c) CNN-LSTM evaluation for multiple past and future horizons.}
\end{figure}

Here we provide the additional graphics mentioned in the main text.
\cref{si-subfig:mae-loss,si-subfig:smape-loss} show the prediction error for all models, when measured using the MAE and SMAPE respectively.
The same conclusions emerge as from the RMSE analysis presented in the main paper.
\cref{si-subfig:CNN-LSTM-best-R-for-P} shows the best value of the past time horizon ($R$) for each future time horizon ($P$) for the hybrid CNN-LSTM.
The graphic resembles closely the graphic for LSTM (\cref{Fig13_14_15}c) discussed in the main text.
%

\end{document}